%% file: paper.tex
\documentclass[10pt,twocolumn,letterpaper]{article}
\pdfoutput=1
\usepackage[utf8]{inputenc}
\usepackage{wacv}
\makeatletter
\@namedef{ver@everyshi.sty}{}
\makeatother
\usepackage{times}
\usepackage{epsfig}
\usepackage{graphicx}
\usepackage{amsmath}
\usepackage{amssymb}
\usepackage{booktabs}
\usepackage{tabularx}
\usepackage{amsmath}
\usepackage{mleftright}
\usepackage[table,dvipsnames]{xcolor}
\usepackage{adjustbox}
\usepackage{multicol}

\usepackage{array}
\usepackage{MnSymbol}
\usepackage{colortbl}
\usepackage{multirow}

\usepackage{tikz}
\usetikzlibrary{shapes.geometric,positioning,shapes.multipart,backgrounds,arrows,patterns,calc}
\usepackage{pgfplots}
\usepackage{subcaption}
\usepackage{array}
\usepackage[absolute,overlay]{textpos}

\usepackage[numbers,sort]{natbib}
\usepackage{multibib}% Needs to be after natbib?!
\newcites{two}{ }% for the CV at the end! It has a separate
\usepackage{etoolbox}

\def\wacvPaperID{1535}
\wacvalgorithmstrack
\wacvfinalcopy

\ifwacvfinal
\usepackage[breaklinks=true,bookmarks=false]{hyperref}
\else
\usepackage[pagebackref=true,colorlinks,bookmarks=false]{hyperref}
\fi

\makeatletter
\pretocmd{\NAT@citexnum}{\@ifnum{\NAT@ctype>\z@}{\let\NAT@hyper@\relax}{}}{}{}
\makeatother

\input{macros}

\begin{document}
\graphicspath{{figures/}}
\title{Learning 3D Human Pose Estimation from Dozens of Datasets using a\\ Geometry-Aware Autoencoder to Bridge Between Skeleton Formats}

\author{István Sárándi \quad Alexander Hermans \quad Bastian Leibe\\
RWTH Aachen University, Germany\\
{\tt\small \{sarandi,hermans,leibe\}@vision.rwth-aachen.de}
}

\maketitle
\thispagestyle{empty}
\begin{textblock*}{21cm}(0mm,15mm)\noindent 
\begin{center}
\textcolor{gray}{2023 IEEE/CVF Winter Conference on Applications of Computer Vision (WACV)}
\end{center}
\end{textblock*}

% Space hacks
%% Caption skips
\setlength{\abovecaptionskip}{3pt plus 4pt minus 2pt}
%% around equations
\setlength{\abovedisplayskip}{6pt}
\setlength{\belowdisplayskip}{6pt}
\setlength\abovedisplayshortskip{6pt}
\setlength\belowdisplayshortskip{6pt}

\aboverulesep = 0.4mm
\belowrulesep = 0.5mm

\setlength{\floatsep}{5pt plus2pt minus2pt}
\setlength{\textfloatsep}{7pt plus3pt minus3pt}
\setlength{\dblfloatsep}{5pt plus2pt minus1pt}
\setlength{\dbltextfloatsep}{7pt plus3pt minus3pt}

\input{body}

{\small
\setlength{\bibsep}{0pt plus 0.3ex}
\bibliographystyle{plainnat}
\bibliography{abbrev_short,paper}
}

%%%%%%%%%%%%%%%% SUPPLEMENTARY

\clearpage

\makeatletter
\newcommand\resetnobreak{\@nobreakfalse}
\twocolumn[\centering \Large \bfseries
Supplementary Material
\resetnobreak\vspace{1cm}]

\setcounter{equation}{0}
\setcounter{figure}{0}
\setcounter{table}{0}
\setcounter{page}{1}
\setcounter{section}{0}

\renewcommand{\theequation}{S\arabic{equation}}
\renewcommand{\thepage}{S\arabic{page}}
\renewcommand{\thesection}{S\arabic{section}}
\renewcommand{\thetable}{S\arabic{table}}
\renewcommand{\thefigure}{S\arabic{figure}}

\input{supp_body}

\end{document}

%% file: macros.tex
\newcommand{\PAR}[1]{\vskip1pt \noindent {\bf #1~}}

\newcommand{\PARbegin}[1]{\noindent {\bf #1~}}

\newcommand{\refsec}[1]{Sec.~\ref{sec:#1}}

\newcommand{\reffig}[1]{Fig.~\ref{fig:#1}}
\newcommand{\reftab}[1]{Tab.~\ref{tab:#1}}

\newcolumntype{Y}{>{\centering\arraybackslash}X}

\newcommand{\f}[1]{\textbf{#1}}

\definecolor{losscolor}{HTML}{E24A33}
\definecolor{netcolor}{HTML}{CCCCCC}
\definecolor{dsa}{HTML}{93b5c6}
\definecolor{dsb}{HTML}{ddedaa}
\definecolor{dsc}{HTML}{f0cf65}
\definecolor{dsd}{HTML}{d7816a}
\definecolor{dse}{HTML}{972544}
\definecolor{latentcolor}{HTML}{51a0cf}
\definecolor{autoencodercolor}{HTML}{a4d0eb}

\newcommand{\sets}{$\diamondsuit$}
\newcommand{\setm}{$\clubsuit$}
\newcommand{\setl}{$\heartsuit$}

%% file: body.tex
\begin{abstract}
Deep learning--based 3D human pose estimation performs best when trained on large amounts of labeled data, making combined learning from many datasets an important research direction.
One obstacle to this endeavor are the different skeleton formats provided by different datasets, \ie, they do not label the same set of anatomical landmarks.
There is little prior research on how to best supervise one model with such discrepant labels.
We show that simply using separate output heads for different skeletons results in inconsistent depth estimates and insufficient information sharing across skeletons.
As a remedy, we propose a novel affine-combining autoencoder (ACAE) method to perform dimensionality reduction on the number of landmarks.
The discovered latent 3D points capture the redundancy among skeletons, enabling enhanced information sharing when used for consistency regularization.
Our approach scales to an extreme multi-dataset regime, where we use 28 3D human pose datasets to supervise one model, which outperforms prior work on a range of benchmarks, including the challenging 3D Poses in the Wild (3DPW) dataset. 
Our code and models are available for research purposes.\footnote{\url{https://vision.rwth-aachen.de/wacv23sarandi}}
\end{abstract}

\begin{figure}[t]
\centering
\includegraphics[width=0.92\linewidth]{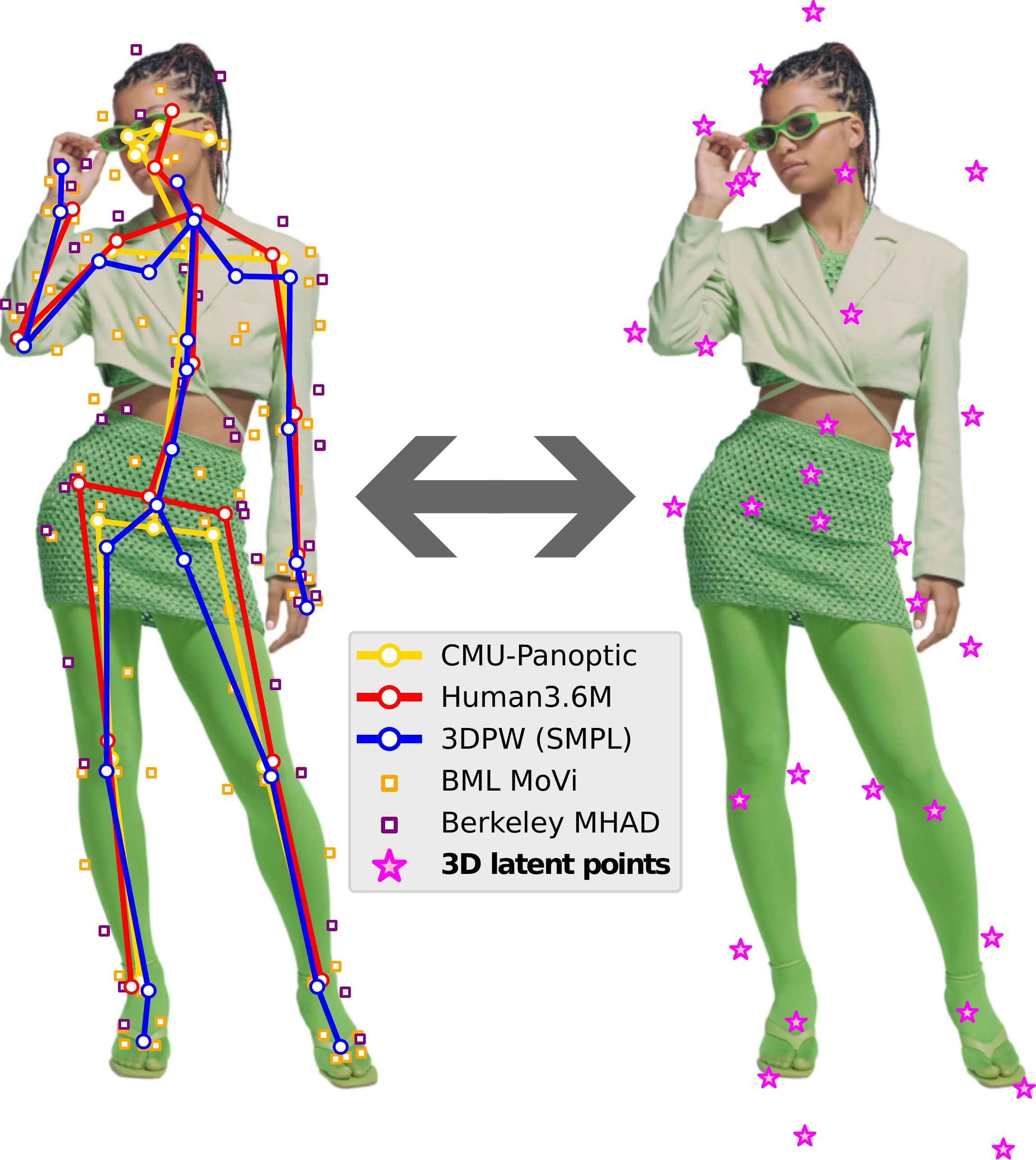}\\
\caption{Different 3D human pose datasets (\eg, CMU-Panoptic and Human3.6M) provide annotations for different sets of body landmarks (\textit{left}).
To best leverage such discrepant labels for multi-dataset 3D pose estimation,
we discover a smaller set of latent 3D keypoints (\textit{right}), from which the dataset-specific points can be reconstructed.
This allows us to capture the redundancy among the different skeleton formats and enhance information sharing between datasets, ultimately leading to improved pose accuracy.}
\label{fig:illust}
\end{figure}

\input{includes/fig_qualitative_results}
\input{includes/tab_datasets}

\vspace*{-7pt}
\section{Introduction}
\vspace*{-2pt}

Research on 3D human pose estimation has gone through enormous progress in recent years~\cite{Sun21ICCV,Joo21ThreeDV,Cheng21CVPR,Liu20CVPR,Moon19ICCV,Fabbri20CVPR,Lin21CVPR,Kocabas19CVPR}.
While semi-supervised and self-supervised approaches are on the rise~\cite{Kundu22CVPR,Zhang21Electronics}, best results are still achieved when using as much labeled training data as possible.
However, individual 3D pose datasets tend to be rather small and lacking in diversity, as they are often recorded in a single studio with few subjects.
Therefore, to provide the best possible models for downstream applications (\eg, action recognition, sports analysis, medical rehabilitation, collaborative robotics), it becomes important to use many datasets in the training process.
Thanks to sustained efforts by the research community, numerous publicly released, labeled datasets exist.
However, as prior published works only train on at most a handful of them, it remains unknown what performance could be achieved by combining more than a decade of dataset collection efforts into a single model.
Unfortunately, this is not a trivial undertaking, since different datasets do not use the same skeleton format for their labels (see \reffig{illust}), \eg, the hip keypoints are at different heights, some body parts are only labeled in some datasets, some provide surface markers while others provide keypoints inside the body, etc.
Prior work has rectified such differences through a handful of individually defined rules (\eg, shrink the hip--pelvis distance by a certain factor~\cite{RapczynskiSensors21}), but this does not scale to many keypoints and datasets---we need a more systematic and automatic method.
The question we tackle in this work is therefore: \textit{How can we automatically merge dozens of 3D pose datasets into one training process, given the label discrepancies?}
We refer to this task as \textit{multi-skeleton 3D human pose estimation}.\footnote{For simplicity, we call any set of landmarks provided in a particular dataset a ``skeleton'', and use ``landmark'' and ``joint'' synonymously.}

If we ignored the discrepancies altogether and proceeded as if keypoints with the same name represented the same body landmark, the model would be supervised with inconsistently labeled examples and would learn to output a skeleton format that is some kind of average of the true ones, leading to subpar benchmark performance.
Alternatively, we may consider this as a multi-task learning problem, and predict the skeletons on separate output heads on a shared backbone, without assuming any skeleton correspondences.
But as we will see, this is not ideal either, as there is insufficient information sharing between skeletons, which is most apparent in inconsistencies between the depth predictions of such a model, as shown in \reffig{qualitative_results}a.

To strike the right balance between those two extremes, we aim to establish \textit{some} connections between the skeleton formats without assuming them to be the same.
To learn such geometric relations between skeletons, we introduce a novel autoencoder-based dimensionality reduction technique to compress a larger set of 3D keypoints (the joints from all datasets) into a lower-cardinality representation (a smaller latent keypoint set).
The encoder and decoder compute affine combinations of their input points, and are thus equivariant to rotation and translation.
We further induce chirality equivariance (left--right symmetry) via weight sharing~\cite{Yeh19NeurIPS}.
We call this model an Affine-Combining Autoencoder (ACAE).
We employ the ACAE in pose estimation training as an output regularizer, to encourage consistent predictions.
This improves prediction results both qualitatively and quantitatively.
As an alternative to the regularization approach, we can also directly predict the latent keypoints of the ACAE with a 3D pose estimator.
This latter variant avoids the need for the underlying pose estimator to estimate a large number of joints, which may be costly for some methods.
In both cases, the final predictions become consistent, showing the value of our approach in tackling multi-dataset 3D pose estimation.

Through an extensive literature review, we have identified 28 datasets with high-quality 3D human pose labels.
By systematically preprocessing these datasets and discarding redundant poses, we constructed a meta-dataset of 13 million 
examples, spanning more than a thousand people.
This is almost two orders of magnitude more data than in typical research papers (\eg, Human3.6M has $\sim$165k examples after redundancy filtering).
We show that using more data indeed helps, and that our approach scales to 28 datasets providing a total of 555 joints in their skeleton formats, summarized in \reftab{datasets}.
Our final models show excellent in-the-wild performance, outperforming currently available models, making them highly useful for downstream research.

In summary, we make the following \textbf{contributions} in this paper.
(1) We assemble the largest scale meta-dataset for 3D human pose estimation to date, consisting of 28 individual datasets, and release scripts for reproducing the process. We call special attention to the problem of disparate skeleton annotation formats in these datasets, which has rarely been addressed in the literature so far.
(2) We propose affine-combining autoencoders (ACAE), a novel linear dimensionality reduction technique applicable to keypoint-based representations such as poses.
(3) We apply the ACAE to regularize model predictions to become more consistent, leading to qualitative and quantitative improvements, and we show that the latent points can be predicted directly as well.
(4) We release high-quality 3D pose estimation models with excellent and consistent in-the-wild performance due to diverse supervision and our regularization tying together different skeleton formats.

\input{figures/training_figure}

\section{Related Work}

\PAR{3D Human Pose Estimation.}
For an overview on the current trends in 3D pose estimator design, we refer the reader to excellent current surveys~\cite{Liu22CSUR, Sang19Tsinghua,Chen20CVIU,Zhang21Electronics,Ji20VRIH}.
We emphasize that our approach is independent of the internals of the pose estimation method.

\PAR{Handling Discrepancy in Skeleton Formats.}
In 2D-to-3D pose lifting, \citet{RapczynskiSensors21} combine pairs of datasets in training by concatenating the training data sets, and harmonize the joints through hand-crafted rules.
In 2D pose tracking, \citet{Guo18ECCVW} train dataset-specific output heads and combine their results via hand-crafted rules.
To unify pose representations, some prior works on (image-independent) MoCap data standardized the height and bone length of skeletons~\cite{Mandery16TOR,Holden16TOG}.
The AMASS dataset~\cite{amass:Mahmood19ICCV} addresses the problem of discrepancy in MoCap data representations by mapping them to the SMPL~\cite{bodymodel:smpl} representation, but the dataset does not provide corresponding images and cannot be used for image-based pose estimation.
Furthermore, the underlying MoSH++ algorithm relies on a complex, multi-stage procedure requiring temporal sequence data and a pre-existing body mesh model.
In contrast, our method has different goals and is much simpler in comparison.
We do not aim to generate a definitive, universal ground truth representation for all datasets, instead our latent keypoint set is only used as an intermediate representation for the pose estimator, but the losses and evaluations are still computed in the original skeleton formats, after decoding the latent points into full skeletons.

\PAR{Keypoint Discovery.}
Discovering a good set of landmarks to describe objects has been investigated in other contexts in computer vision.
2D keypoint discovery has been used to disentangle pose and shape in 2D human pose estimation~\cite{Jakab18NeurIPS,Jakab20CVPR}. 
In 3D, \citet{Jakab21CVPR} discover control points for deforming 3D shapes.
\citet{Rhodin18ECCV} learn a 3D human representation that consists of a set of 3D points, which encode both pose and appearance, optimizing for the unsupervised auxiliary task of novel view synthesis.
\citet{mosh:Loper14TOG} optimize the placement of sparse markers on the body to best capture both human shape and pose.

\PAR{Linear Subspace Learning.}
Linear dimensionality reduction has a long history, with principal component analysis being the best known representative~\cite{Pearson01}. 
Its relation to autoencoders was discovered by Bourlard and Kamp~\cite{Bourlard88BC}, and a recent paper by the same first author reviews the developments since~\cite{Bourlard22BC}.
Linear autoencoders have been employed in robust and sparse~\cite{Guerra21Psychometrika} variants, a detailed overview is presented in \cite{Cunningham15JMLR}.
Our proposed affine-combining autoencoders are related, but have different constraints, tailored to our use case, \ie, that the weights sum to unity, and there is no requirement of orthogonality, unlike in PCA.

\section{Method}
\label{sec:method}
\vspace*{-2pt}
Our goal is to obtain a strong, monocular RGB-based 3D human pose estimation model by integrating numerous datasets into one mixed training process, even when the different datasets provide annotations according to different skeleton formats.
Suppose we have $D$ skeleton formats, with $\{J_d\}_{d=1}^{D}$ joints in each, for a total of $J=\sum_{d=1}^{D}J_d$ joints overall.
Further, we have a merged dataset with $N$ training examples, each consisting of an image of a person and annotations for a subset of the $J$ body joints in 3D.
Our proposed workflow consists of three main steps.
First, we train an initial model that predicts the different skeletons on separate prediction heads, branching out from a common backbone network (\reffig{train_init}).
With the resulting model, we can run inference and produce a pseudo--ground truth ``parallel corpus'' of many poses given in every skeleton format.
From this, the geometric relations between skeleton formats can be captured.
We accomplish this in the second step, by training an undercomplete geometry-aware autoencoder, which discovers a latent 3D body landmark set that best captures human pose variations in the pseudo-GT data (\reffig{train_ae}).
Finally, equipped with the trained autoencoder, we rely on its learned latent space to make the model output consistent across skeleton formats through output regularization (\reffig{tune_with_reg}).
We also experiment with direct latent point prediction, and a hybrid variant for the last step.

\subsection{Initial Model Training}
\label{sec:initmodel}
\vspace*{-2pt}
The first step of our workflow is to train an initial pose estimator to predict all $J$ joints separately (\reffig{train_init}).
This means that no correspondences or relations across different skeletons are assumed, \ie, without specifying or enforcing that the left shoulder joint of one skeleton should be predicted near the left shoulder of another skeleton.
This is akin to multi-task architectures that use different task-specific heads on one backbone.
The pose loss we minimize is $\mathcal{L}_{\text{pose}} = \mathcal{L}_{\text{meanrel}} + \lambda_{\text{proj}}\mathcal{L}_{\text{proj}} + \lambda_{\text{abs}}\mathcal{L}_{\text{abs}}$, where $\mathcal{L}_{\text{meanrel}}$ is an $\ell_1$ loss computed after aligning the prediction and ground truth at the mean, $\mathcal{L}_{\text{proj}}$ is an $\ell_1$ loss on the 2D coordinates after projection onto the image, and $\mathcal{L}_{\text{abs}}$ is an $\ell_1$ loss on the absolute pose (in camera coordinates).
Since each training example is annotated only with a subset of the $J$ joints, we ignore any unlabeled joints when averaging the loss.

When visualizing the different skeleton outputs of this trained model, we see inconsistencies among them along the challenging depth axis (see \reffig{qualitative_results}).
This is understandable, since we have not employed any training mechanism that would ensure any relations between the output skeletons (except that they are predicted from shared backbone features).
On the other hand, when projected onto the image plane, the predictions appear sufficiently consistent.

\subsection{Pseudo--Ground Truth Generation}
\vspace*{-2pt}
To characterize how the joints of the different skeletons relate to one another, we need pose labels according to all skeleton formats for the same examples, to function as a ``Rosetta Stone''.
Since no such ground truth is available (datasets only provide one type of skeletons, rarely two), we generate pseudo--ground truth using the initial separate-head model.
It is important to use images that the model can handle well in this step, hence we choose a relatively clean, clutter-free subset of the \textit{training} data for this purpose (H36M and MoVi).
This yields a set of $K$ pseudo--ground truth poses, with all $J$ joints: $\left\{ P_k \in \mathbb{R}^{J\times 3} \right\}_{k=1}^{K}$.

\subsection{Affine-Combining Autoencoder}
\vspace*{-2pt}
To capture the redundancy among the full set of $J$ joints, and ultimately to improve the consistency in estimating them, we introduce a simple but effective dimensionality reduction technique.
Since the pseudo-GT is more reliable in 2D (the X and Y axes) than in the depth dimension, the transformation to and from the latent representation should be viewpoint-independent, in other words the representation should be equivariant to rotation and translation.
This equivariance in turn requires the latent representation to be geometric, \ie, to consist of a list of $L$ latent 3D points $Q_k \in \mathbb{R}^{L\times 3}$ ($L < J$).
This makes intuitive sense: the way the different skeletons relate to each other is only dependent on how joints are defined on the human body, not on the camera angle.
The latent points are then responsible for spanning the overall structure of a pose.
Specific skeleton formats can then be computed in relation to these latents.
Further, the latent points should only have sparse influence on the joints, \eg, some latent points should be responsible for the positioning of the left arm and these should have no influence on the right leg's pose.
We find that these requirements can be fulfilled effectively by adopting a novel constrained undercomplete linear autoencoder structure, which we call \textit{affine-combining autoencoder} (ACAE).
Instead of operating on general $n$-dimensional vectors, an ACAE's \textbf{encoder} takes as input a list of $J$ points $\mathbf{p}_j \in \mathbb{R}^3$ and encodes them into $L$ latent points $\mathbf{q}_l \in \mathbb{R}^3$ by computing affine combinations according to%
\begin{equation}
\mathbf{q}_l = \sum_{j=1}^{J} w_{l,j}^{\text{enc}}\mathbf{p}_j, \quad \sum_{j=1}^{J} w_{l,j}^{\text{enc}} = 1, \quad \forall l=1, \ldots, L. \\
\end{equation}

Similarly, the \textbf{decoder}'s goal is to reproduce the original points from the latents, again through affine combinations:%
\begin{equation}
\mathbf{\hat{p}}_j = \sum_{l=1}^{L} w_{j,l}^{\text{dec}} \mathbf{q}_l, \quad \sum_{l=1}^{L} w_{j,l}^{\text{dec}} = 1, \quad \forall j=1, \ldots, J. \\
\end{equation}

Since affine combinations are equivariant to any affine transformation, our encoder and decoder are guaranteed to be rotation and translation equivariant.
(Note that the same weighting is used for the X, Y and Z coordinates.)

The learnable parameters of the ACAE are the affine combination weights $w_{l,j}^{\text{enc}}$ and $w_{j,l}^{\text{dec}}$, which can also be understood as (potentially negative) generalized barycentric coordinates~\cite{Hormann17CRC} for the latents \wrt the full joint set and vice versa.
Allowing negative coordinates is necessary, as this allows the latents to spread outwards from the body, similar to a cage used in graphics~\cite{Nieto2013}.
Restricting the encoder and decoder to convex combinations would severely limit its expressiveness.
To achieve sparsity in the weights (\ie, spatially localized influence), we use $\ell_1$ regularization, and this also reduces the amount of negative weights, preferring nearly convex combinations.
We further adopt the $\ell_1$ reconstruction loss, as it is robust to outliers which may be present due to noise in the pseudo-GT.

\PAR{Problem Statement.} We can now formally state our proposed ACAE problem in matrix notation for the weights. Given $K$ training poses with $J$ joints $\left\{ P_k \in \mathbb{R}^{J\times 3} \right\}_{k=1}^{K}$,%
\begin{align}
\label{eq:autoencoder}
\begin{split}
\underset{W_{\text{enc}}\in \mathbb{R}^{L\times J},\, W_{\text{dec}}\in \mathbb{R}^{J\times L}}{\text{minimize}}\ & \mathcal{L}_{\text{reconstr}} + \lambda_{\text{sparse}} \mathcal{L}_{\text{sparse}} \\
 \mathcal{L}_{\text{reconstr}} &= \frac{1}{K}\sum_{k=1}^{K} \left\| P_k - W_{\text{dec}} W_{\text{enc}} P_k  \right\|_1 \\
 \mathcal{L}_{\text{sparse}} &= \left\|W_{\text{enc}}\right\|_1 + \left\|W_{\text{dec}}\right\|_1 \\
\text{s. t.}\quad W_{\text{enc}}\mathbf{1}_{J} &= \mathbf{1}_{L},\quad W_{\text{dec}}\mathbf{1}_{L} = \mathbf{1}_{J},
\end{split}
\end{align}
where $\mathbf{1}_{a}$ is a vector of dimension $a$ filled with ones and $\lambda_{\text{sparse}}$ controls the strength of the sparsity regularization.
The sum-to-one (partitioning of unity) constraints ensure that the weights express affine combinations, which is necessary for translation equivariance.

\PAR{Reconstruction Loss on 2D Projection.} As discussed above, the pseudo-GT is more reliable in its 2D projection than along the depth axis.
We therefore adapt the above general problem formulation to take this into account by defining the reconstruction loss on 2D projections:%
\begin{align}%
\mathcal{L}_{\text{reconstr}}^{\text{proj}} &= \frac{1}{K}\sum_{k=1}^{K} \left\| \Pi\left(P_k\right) - \Pi\left(W_{\text{dec}} W_{\text{enc}} P_k\right)  \right\|_1,
\end{align}
where $\Pi(\cdot)$ denotes camera projection.

Our key insight here is that it is sufficient to observe the high-quality 2D image-plane projections of this model's outputs to characterize how the joints of different skeleton formats geometrically interrelate, because these relations are viewpoint-independent.
As a simplified example, if we observe on many poses, that a certain joint tends to be halfway in between two other joints in 2D, then this will also have to hold along the depth axis.

\PAR{Chirality Equivariance.} 
As humans have bilateral symmetry, it is natural to expect the autoencoder to be chirality-equivariant, \ie, to process the left and right sides the same way~\cite{Yeh19NeurIPS}.
To this end, we partition the latent keypoints into three disjoint sets: left, right and central latents, following the same proportions as in the full joint set.
Assume, without loss of generality, that the points are sorted and grouped into left-side, right-side and center blocks.
We then impose the following weight-sharing block structure on both the encoder and decoder weight matrices:%
\begin{equation}
\renewcommand\arraystretch{1.3}
W =
\mleft[\begin{array}{c|c|c}
W_1 & W_2 & W_3 \\
\hline
W_2 & W_1 & W_3 \\
\hline
W_4 & W_4 & W_5
\end{array}\mright].
\end{equation}
This structure indeed ensures chirality equivariance, since the matrix remains the same if we permute both its rows and columns by swapping the first two sections, \ie, swapping the left and right points in the inputs and the outputs.

\PAR{Head Keypoint Weighting.}
Based on the intuition that smaller motions of head and facial keypoints can be more semantically relevant, we weight these joints higher (by a factor of 10) in the loss, ensuring that the latents sufficiently cover the head as well.
(We later found that this is not strictly necessary and the method also works without this as well.)

\PAR{Training.}
We train the autoencoder using the Adam optimizer~\cite{Kingma15ICLR} with batch size 32.
To enforce the sum-to-one constraints, we normalize the weight matrices within the computational graph.

\subsection{Consistency Fine-Tuning}
\label{sec:finetuning}
\vspace*{-2pt}
Once our affine-combining autoencoder is trained on pseudo--ground truth, we freeze its weights and use it to enhance the consistency of 3D pose estimation outputs, with one of three alternative methods.

\input{figures/latent_and_latentpred_model}

\PAR{Output Regularization.}
In this case (\reffig{tune_with_reg}), we estimate all $J$ joints $\hat{P} \in \mathbb{R}^{J\times 3}$ with the underlying pose estimator, but we feed this output through the autoencoder, and apply an additional loss term that measures the consistency of the prediction with the latent space, through an $\ell_1$ loss, as%
\begin{align}%
\mathcal{L}_{\text{cons}} &= \left\| \hat{P} - W_{\text{dec}} W_{\text{enc}} \hat{P} \right\|_1.
\end{align}
This encourages that the separately predicted skeletons can be projected to latent keypoints and back without information loss, thereby discouraging inconsistencies between them.
The pose loss $\mathcal{L}_{\text{pose}}$ (\cf \refsec{initmodel}) is applied on $\hat{P}$.

\PAR{Direct Latent Prediction.}
To avoid having to predict a large number of $J$ joints in the base pose estimator, we define an alternative approach where the latents $\hat{Q} \in \mathbb{R}^{L\times 3}$ are directly predicted and then fed to the frozen decoder (\reffig{latent_prediction_arch}).
The last layer is reinitialized from scratch, as the number of predicted joints changes from $J$ to $L$.
The pose loss $\mathcal{L}_{\text{pose}}$ is applied on $W_{\text{dec}} \hat{Q}$.

\PAR{Hybrid Student--Teacher.}
In a hybrid of the above two variants, we keep the full prediction head and add a newly initialized one to predict the latents $\hat{Q}$ directly (\reffig{student_teacher_arch}).
To distill the knowledge of the full prediction head to the latent head, we add a student--teacher-like $\ell_1$ loss%
\begin{align}%
\mathcal{L}_{\text{teach}} &= \left\| \hat{Q} - \texttt{stop\_gradient}\left(W_{\text{enc}} \hat{P}\right) \right\|_1,
\end{align}
which is only backpropagated to the latent predictor (the student).
During inference, we use $W_{\text{dec}} \hat{Q}$ as the output, to be as lightweight as direct latent prediction.

\section{Experimental Setup}
\vspace*{-2pt}

\PAR{Base Model.}
We adopt the recent state-of-the-art MeTRAbs~\cite{Sarandi21TBIOM} 3D human pose estimator as the platform for our experiments, but we note that our method is agnostic to the specifics of the underlying pose estimator.
Unless mentioned otherwise, the backbone is EfficientNetV2-S~\cite{Tan21ICML}.

\PAR{Training Details.}
We perform 400k training steps with AdamW~\cite{Loshchilov19ICLR} and batch size 128, with every dataset represented with a fixed number of examples per batch.
Batch composition and learning rate schedule are specified in the supplementary.
We use ghost BatchNorm~\cite{Hoffer17NIPS,Summers19ICLR} with size 16, as this improved convergence in multi-dataset training, together with switching the BatchNorm layers to inference mode for the last 1000 updates.
The final fine-tuning phase has 40k iterations with smaller learning rate on the backbone than the heads.
The autoencoder weights are trained on pseudo-GT obtained with EffV2-L.
We crop a 256$\times$256 px square around the person, apply perspective undistortion with camera intrinsics and perform augmentation as in~\cite{Sarandi21TBIOM}.

\PAR{Datasets.}
See \reftab{datasets} for an overview of all used datasets, which employ a variety of skeleton formats.
In some cases, \eg, when annotations are derived through triangulating COCO-like predictions (of \eg, OpenPose), or through fitting a body model (\eg, SMPL), we can assume that multiple datasets use the same convention (indicated in the last column).
For other datasets, we assume the skeleton is a custom one, yielding 555 distinct keypoints in total.
As most 3D human datasets contain videos, rather than isolated images, the number of sufficiently different poses is smaller than the total number of annotated frames.
We hence discard examples where all joints remain within 100~mm of the last stored example.
Our overall processing ensures that each training example has a person-centered image crop, camera intrinsics, 3D coordinates for some subset of the joints, a bounding box and a segmentation mask.

\input{includes/tab_data_sizes}
\input{includes/tab_main}

\PAR{Evaluation Metrics.}
We evaluate on four datasets: MuPoTS~\cite{ds:mucomupots}, 3DPW~\cite{ds:3dpw}, 3DHP~\cite{ds:3dhp} and H36M~\cite{ds:h36m1,ds:h36m2}.
Over the years, different evaluation metrics and protocols have become customary on different datasets, whose details can be very arcane.
Especially in a multi-dataset setting, we find it important to use consistent metrics.
For our main experiments, we therefore adopt the following four metrics everywhere:
MPJPE: mean Euclidean distance between predicted and ground truth joints after alignment at the root joint.
PMPJPE: mean Euclidean distance after Procrustes alignment. PCK@100mm: percentage of joints predicted within 100~mm of the ground truth after root alignment. CPS@200mm: percentage of poses where \emph{all} joints are within 200~mm distance of the ground truth after root alignment~\cite{Wandt21CVPR}.
More details on the experimental setup are in the supplementary.

\section{Results}
\vspace*{-2pt}
\subsection{Benefit of Training on Many Datasets}
\vspace*{-2pt}
Since one contribution of our paper is the study of the large-scale multi-dataset training regime, an important question is whether this brings improvements or whether performance saturates with just a few large-scale datasets.
As a simple baseline, we train models on individual datasets and evaluate on the corresponding test splits.
(With MuPoTS, we use MuCo-3DHP for training).
We then train on three dataset combinations, as shown in \reftab{dataset_sizes}.
There is a clear trend showing performance improvement when training with more datasets, and the small dataset combination also outperforms single-dataset baselines.
We note that H36M scores sometimes suffer from additional data.
H36M uses the same studio environment in the training and evaluation split, therefore the model works better when a large part of the training batches are filled with H36M examples, allowing it to specialize on images from this room, but this does not reflect true generalization ability.
The model trained on the large dataset combination achieves very strong scores across the board, confirming that using many datasets makes a difference.

Despite the good benchmark scores, we qualitatively observe (\reffig{qualitative_results}) that the different skeleton outputs can still be inconsistent among themselves.

\subsection{Consistent Multi-Skeleton Prediction}
\vspace*{-2pt}
A first naive baseline for achieving consistent predictions is to merge joints from different skeletons (\eg, we predict only one ``left shoulder'' joint), reducing the joint count from 555 to 163.
This leads to weaker results than predicting all joints separately (see \reftab{approach_comparisons}), since joints with similar names may represent somewhat different keypoints.

H36M is again an outlier, as the prediction with merged joints works well for it.
Since the model can easily recognize that a test image comes from the H36M studio, it can adapt its prediction to match the H36M skeleton format.
This is not possible on \eg, 3DPW, since the model cannot know in advance what skeleton format will be used for the reference poses of these images, since they come from diverse in-the-wild scenes.

When using our proposed ACAE-based regularization (\cf \reffig{tune_with_reg}), we can see consistent improvements for almost all metrics.
However, the improvement in the qualitative performance of the model is even more striking.
As seen in \reffig{qualitative_results}, the regularized model creates significantly more consistent skeleton predictions.
Especially the depth-consistency is improved, but some errors in the frontal view are also corrected.
More qualitative results in the supplementary show that this observation holds broadly.

Overall, the model that estimates latent keypoints (\reffig{latent_prediction_arch}) has slightly lower performance than the separate-head baseline, likely because latent keypoints may be placed at less characteristic locations on the body and can thus be harder to localize.
Further, the latent keypoint head's weights are initialized from scratch, whereas the regularization-based method fine-tunes a pretrained head.
The hybrid combination from \reffig{latent_prediction_arch} performs slightly worse than the model that is only regularized, but in many cases still outperforms the baseline.
This shows that a direct estimation of the discovered latent keypoints is also a viable option.
By design, this approach also produces consistent results, since we compute a single latent set of keypoints from which we decode all skeletons.

We also train the regularization and hybrid variants with EffNetV2-Large (lower part of \reftab{approach_comparisons}).
Overall, the results follow the same order, and they are better across the board.
Regularization improves results and also leads to consistent predictions, and the hybrid approach is somewhat better than the initial model trained to predict separate joints.

This means that our autoencoder-based regularization is effective at improving results both quantitatively and qualitatively, and the discovered latent keypoints can be predicted directly.
This opens up interesting future research directions, as the latent keypoints can be seen as a model agnostic interface, potentially allowing us to incorporate new skeleton formats by expanding the decoder, without a need for model specific fine-tuning or probing.

\subsection{Comparison to Prior Works}
\vspace*{-2pt}
\input{includes/tab_sota}
In \reftab{sota}, we compare our final results to recent state-of-the-art published works (using standard protocols) and observe much better accuracy than SOTA models.
We emphasize that this comparison is not ``fair'' \wrt the amount of training data.
However, our goal in this paper is to show the value in large-scale multi-dataset training, and to investigate how to best supervise models in that setting.

\subsection{Ablations}
\vspace*{-2pt}

\PARbegin{Chirality Equivariance Constraints.} In \reftab{chirality_effect}, we analyze the effect of enforcing chirality equivariance on the ACAE.
In the quantitative metrics, we see approximately no change or a slight positive effect on both evaluated models.
Given that symmetry makes sense as an inductive bias, we use chirality equivariance in our default setting.

\PAR{Latent Keypoint Count.} \reffig{dimensionality} shows that once a minimum number of latent points is reached, the reconstruction error only decreases slowly (evaluated on a held-out pseudo-GT validation set).
We evaluate several latent sizes for fine-tuning in Tab.~\ref{tab:latent_size}.
48 points work well in practice, and our regularization method is robust \wrt this hyperparameter.
When directly predicting latent keypoints, using too few or too many latent keypoints has a negative effect, but the differences are small beyond 32.
\input{includes/tab_chirality_effect}
\input{includes/tab_latent_size}

\begin{figure}[t]
\centering
\begin{tikzpicture}[tight background]
\begin{axis}[ 
    width=0.95\linewidth,
    height=3.2cm,
    grid=major, % Display a grid
    tick label style={font=\scriptsize},
    label style={font=\scriptsize},
    xlabel={\# Latent keypoints},
    ylabel style={align=center},
    ylabel={Mean projected \\ joint error (mm)},
    xtick={32,64,128,256,512,555},
    % xmode=log,
    ytick={0,5,10,15,20,25,30},
    y tick label style={
        /pgf/number format/.cd,
        fixed,
        precision=0
    },
    xlabel shift = -4pt,
    ylabel shift = -4pt,
    xmin=0, xmax=555,
    ymin=0, ymax=32,
]
\addplot[no marks, color=latentcolor, line width=1] table [x expr=\thisrowno{0}, y expr=\thisrowno{1}, col sep=comma] {figures/latent_dimensionality_plot.csv};
\end{axis}%
\end{tikzpicture}%
\caption{Intrinsic dimension analysis of our pseudo--ground truth with 555 joints.
The residual error curve shows a characteristic elbow shape.}
\label{fig:dimensionality}
\end{figure}

\section{Conclusion}
\vspace*{-2pt}
We have proposed a principled, automatic approach to the problem of large-scale multi-skeleton training of 3D human pose estimation.
Despite its practical relevance in exploiting a large number of 3D pose datasets in one training, this problem has been largely overlooked in the literature.

Our approach relies on a novel formulation of dimensionality reduction of sets of keypoints, via an affine-combining autoencoder with guaranteed built-in equivariances to common transformations.
By regularizing a 3D human pose estimator's output to stay close to the learned latent space discovered by the autoencoder, we can more effectively share information between the different datasets, resulting in an overall more accurate and consistent pose estimator.
We release code for data processing and training, as well as trained models to serve as high-quality off-the-shelf methods for downstream research.

\PAR{Acknowledgments.} This work was supported by the ERC Consolidator Grant project ``DeeViSe'' (ERC-CoG-2017-773161) and by Robert Bosch GmbH under the project ``Context Understanding for Autonomous Systems.''

%% file: includes/fig_qualitative_results.tex
\begin{figure*}%
\newcommand{\cropim}[1]{\includegraphics[width=0.165\textwidth,trim={1.9cm 1.9cm 1.6cm 1.7cm},clip]{figures/qual_sep_reg/#1.pdf}}
\newcommand{\figline}[1]{%
    \includegraphics[width=0.24\textwidth]{qual_sep_reg/#1_im.pdf}&&%
    \cropim{#1_sep1}&\cropim{#1_sep2}&&\cropim{#1_reg1}&\cropim{#1_reg2}\\%
}
\setlength{\tabcolsep}{0.0pt}
\scriptsize
\begin{tabularx}{\textwidth}{cp{5pt}YYp{5pt}YY}%
 && \multicolumn{2}{c}{(a) Separate skeleton prediction} && \multicolumn{2}{c}{(b) With our proposed ACAE regularization}\\
&& Front view & Right side view && Front view & Right side view \\
\figline{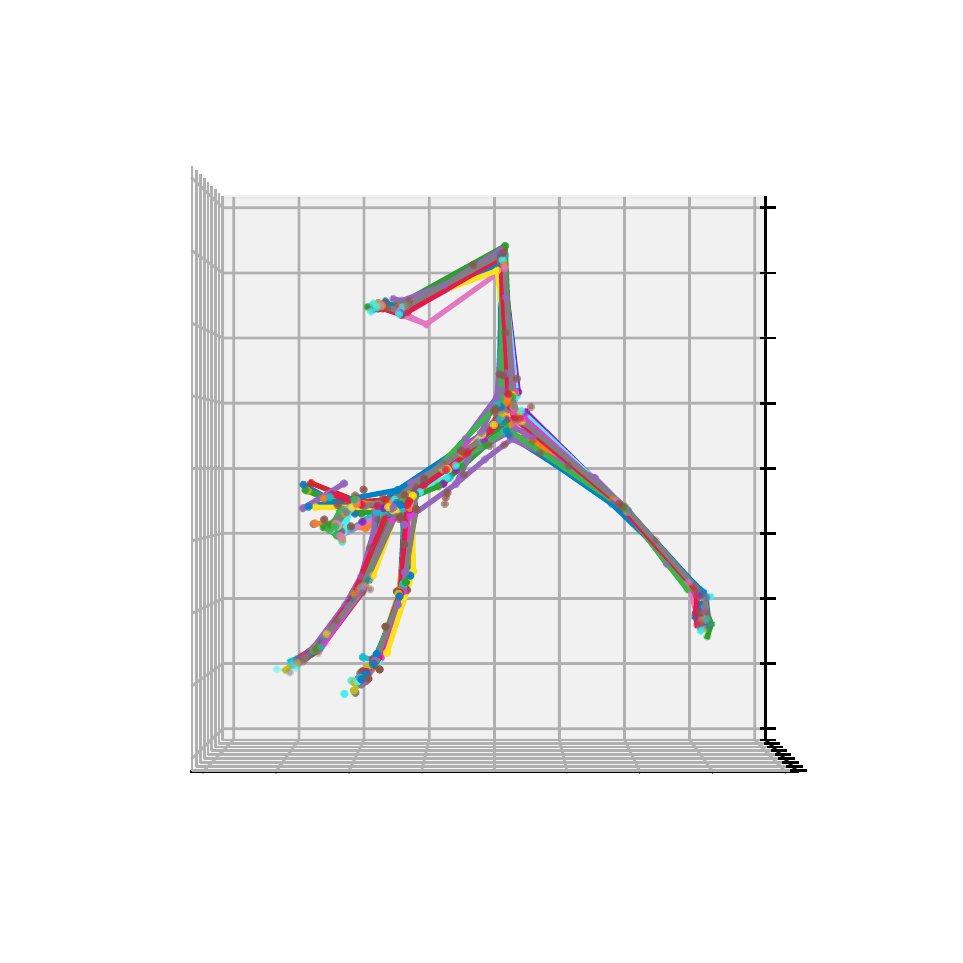}
\figline{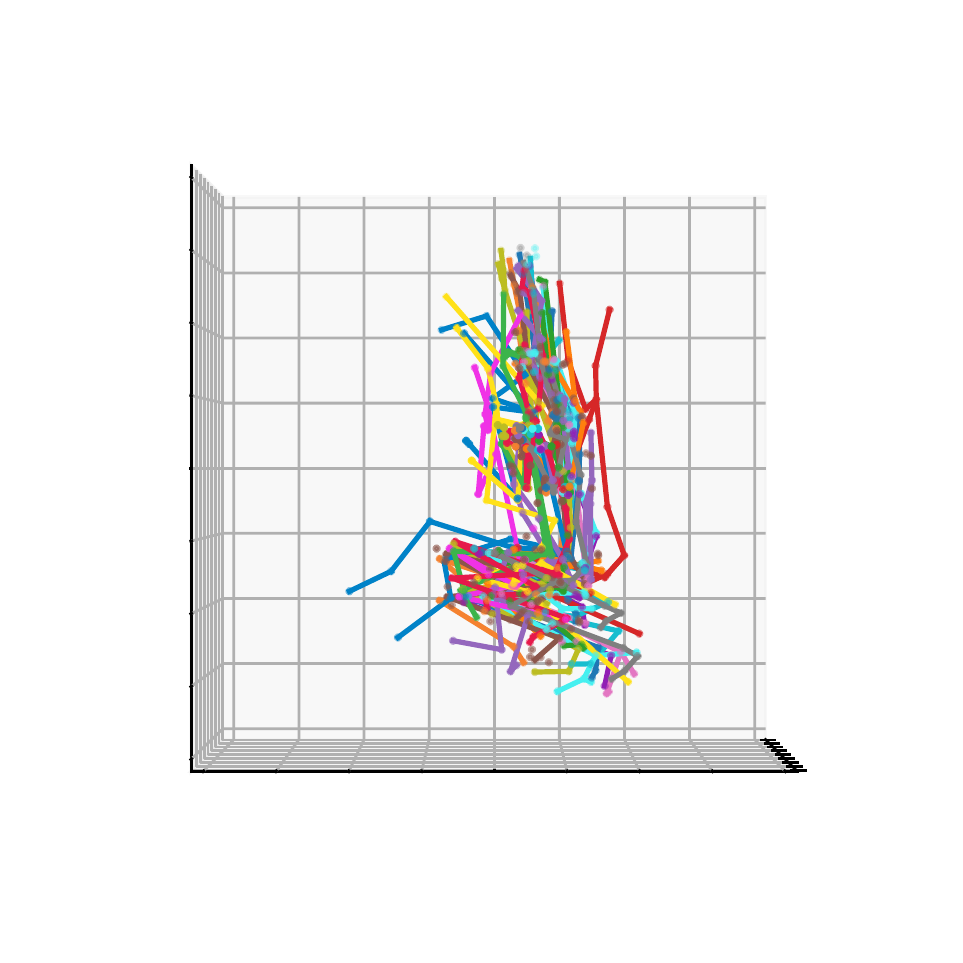}
\end{tabularx}%
\caption{We train models to jointly estimate 3D human pose according to multiple different skeleton formats so that we can train on many datasets at once.
a) Simply using separate prediction heads on a shared backbone is an insufficient solution to this multi-skeleton learning problem, as we obtain inconsistent outputs along the depth axis.
b) We propose a method to capture and exploit the redundancy among the different skeletons using a novel affine-combining autoencoder-based (ACAE) regularization.
This leads to a clear improvement in skeleton consistency.}
\label{fig:qualitative_results}
\end{figure*}

%% file: includes/tab_datasets.tex
\newcolumntype{R}[2]{%
    >{\adjustbox{angle=#1,lap=\width-(#2)}\bgroup}%
    l%
    <{\egroup}%
}
\newcommand*\rot{\multicolumn{1}{R{45}{1em}}}% no optional argument here, please!
\newcommand{\gb}{\rowcolor{gray!10}}
\newcommand{\st}{$\medstar$}

\begin{table}[t]
\caption{
We study the extreme multi-dataset setting of 3D human pose estimation, using all datasets below in one training process.
We define three dataset combinations (indicated by \sets, \setm, and, \setl) to study the effect of training data amount.
(* marks custom dataset-specific skeletons.)
}
\setlength\tabcolsep{1.3mm}
\renewcommand{\arraystretch}{0.932}
\footnotesize
\centering
\begin{tabularx}{\linewidth}{lcYYcc}
\toprule
Dataset name & \rot{\#Examples} & \rot{\#Real Subj.} & \rot{\#Keypoints} & \rot{Subset} &  Skeleton \\
\midrule
\multicolumn{6}{c}{\textit{Real images with markerless MoCap}} \vspace{0.9mm}\\
\gb MuCo-3DHP~\cite{ds:mucomupots} &  677k & 8 & 28 & \sets\setm\setl & 3DHP \\
CMU-Panoptic~\cite{ds:cmupanoptic} &  2.81M & $>$60 & 19 & \setm\setl & COCO  \\
\gb AIST-Dance++~\cite{ds:aist,ds:aistpp} &  1.86M & 30 & 19 & \setm\setl & COCO \\ %,SMPL
HUMBI~\cite{ds:humbi} &  1.26M & 772 & 19 & \setm\setl & COCO \\ %, SMPL
\gb MPI-INF-3DHP~\cite{ds:3dhp} &  627k & 8 & 28 & \setm\setl & 3DHP \\
RICH~\cite{ds:rich} &  96k & 15 & 42  & \setm\setl & SMPL-X~\cite{bodymodel:smplx} \\
\gb BEHAVE~\cite{ds:behave} &  42k & 7 & 43  & \setl & SMPL, COCO \\
ASPset~\cite{ds:aspset} & 124k & 15 & 17 & \setl & *\\
\gb 3DOH50K~\cite{ds:3doh50k} &  50k & $<$10 & 14 & \setl & LSP~\cite{ds:lsp} \\ %, SMPL (24)\\
IKEA ASM~\cite{ds:ikea} &  23k & 48 & 17 & \setl & *\\
\midrule
\multicolumn{6}{c}{\textit{Real images with marker-based MoCap}}\vspace{0.9mm}\\
\gb Human3.6M~\cite{ds:h36m1,ds:h36m2} &  165k & 5 & 25 & \sets\setm\setl & H3.6M \\%Vicon
TotalCapture~\cite{ds:totalcapture} &  130k & 5 & 21 & \setm\setl & *\\%Vicon
\gb BML-MoVi~\cite{ds:bmlmovi} &  553k & 13 & 87 & \setl & * \\ %Qualisys+IMU
Berkeley-MHAD~\cite{ds:bmhad} &  526k & 12 & 43 & \setl & * \\%Vicon
\gb UMPM~\cite{ds:umpm} &  164k & 30 & 15 & \setl &  *\\%Vicon
Fit3D~\cite{ds:fit3d} &  147k & 8 & 25 & \setl & H3.6M \\%Vicon
\gb GPA~\cite{ds:gpa} &  109k & 13 & 34 & \setl & *\\ %Vicon
HumanSC3D~\cite{ds:humansc3d} &  72k & 4 & 25 & \setl & H3.6M\\%Vicon
\gb CHI3D~\cite{ds:chi3d} &  46k & 6 & 25 & \setl & H3.6M \\%Vicon
Human4D~\cite{ds:human4d} &  40k & 4 & 32 & \setl & *\\%
\gb MADS~\cite{ds:mads} &  33k & 5 & 15 & \setl & * \\%Vicon
\midrule
\multicolumn{6}{c}{\textit{Synthetic images}}\vspace{0.9mm}\\
\gb SURREAL~\cite{ds:surreal} &  1.9M & & 24  & \sets\setm\setl & SMPL~\cite{bodymodel:smpl} \\
3DPeople~\cite{ds:3dpeople} &  946k &  & 29  & \setm\setl & * \\
\gb JTA~\cite{ds:jta} &  562k &  & 22 & \setm\setl & * \\
HSPACE~\cite{ds:hspace} &  195k & & 35  & \setm\setl & GHUM~\cite{bodymodel:ghum}  \\
\gb SAIL-VOS~\cite{ds:sailvos} &  101k &  & 26  & \setm\setl & * \\
AGORA~\cite{ds:agora} &  79k &  & 66  & \setm\setl & SMPL[-X] \\
\gb SPEC~\cite{ds:spec} &  59k &  & 24  & \setl & SMPL \\
\midrule
\multicolumn{6}{c}{\textit{Real images with 2D annotations (weak supervision)}}\vspace{0.9mm}\\
\gb COCO~\cite{ds:coco} &  47k &  &  17  &  & \\
MPII~\cite{ds:mpii} &  27k &  & 16  & & \\
\gb PoseTrack~\cite{ds:posetrack} &  40k &  & 15  &  & \\
JRDB~\cite{ds:jrdb} &  59k &  & 17 &  & \\
\midrule
\multicolumn{6}{c}{\textit{Totals (for 3D-labeled data)}}\vspace{0.9mm}\\
\gb Small (3 datasets) & 2.8M & 13  & 77 & \sets & \\
Medium (14 datasets) & 10.8M & $>$900 & 277 & \setm & \\
\gb GRAND TOTAL (28 ds.) & 13.4M & $>$1k & 555 & \setl &  \\
\bottomrule
\end{tabularx}
\label{tab:datasets}
\end{table}

%% file: figures/training_figure.tex
\tikzset{posestyle/.style={rectangle split,rectangle split parts=12, draw, inner sep=0pt, outer sep=0pt,rectangle split part fill={dsa,dsa,dsa,dsb,dsb,dsb, dsc,dsc, dsd,dse}}}
\begin{figure*}
\setlength\tabcolsep{0.4mm}
\begin{tabular}{c|c|c}
\subcaptionbox{Step 1: Train initial model.\label{fig:train_init}}{
\begin{tikzpicture}[font=\small]
    \node (image) [rectangle, draw, minimum width=12mm, minimum height=10mm]{Image};
    \node (net) [right=3mm of image.east, anchor=south, trapezium, trapezium angle=80, minimum width=20mm, minimum height=8mm, draw=netcolor!50!black, thick, rotate=-90,fill=netcolor] {};
    \node (nettext) [anchor=center,align=center]  at (net.center)  {3D Pose \\ Estimator};
    \path (image) edge[-latex] (net);
    \node (poses) [posestyle,right=3mm of net.north, anchor=west, ] {\nodepart{one}\nodepart{two}\nodepart{three}\nodepart{four}\nodepart{five}\nodepart{six}};
    \node[right= 1mm of poses.one west] {\tiny left\_shoulder\_H36M};
    \node[right= 1mm of poses.two west] {\tiny right\_shoulder\_H36M};
    \node[right= 1mm of poses.three west] {\scalebox{0.5}{$\vdots$}};
    \node[right= 1mm of poses.four west] {\tiny left\_shoulder\_3DHP};
    \node[right= 1mm of poses.five west] {\scalebox{0.5}{$\vdots$}};
    \path (net) edge[-latex] (poses);
    \node (loss) [below right= 13mm and 10mm of poses.center, anchor=center, losscolor]{$\mathcal{L}_{\text{pose}}$};
    \draw [-latex,losscolor,dashed] (poses.south) to [out=270,in=180] (loss.west);
\end{tikzpicture}}
&
\hfill
\subcaptionbox{Step 2: Train the autoencoder.\label{fig:train_ae}}{
\hspace*{7pt}
\begin{tikzpicture}[font=\small]
    \node (poses) [posestyle] {};
    \node (enc) [right=3mm of poses.east, anchor=south, trapezium, trapezium angle=65, minimum width=20mm, minimum height=6mm, draw=autoencodercolor!50!black, thick, rotate=-90, fill=autoencodercolor] {};
    \path (poses) edge[-latex] (enc);
    \node (enctext) [anchor=center]  at (enc.center)  {$W_{\text{enc}}$};
    \node (latent) [right=3mm of enc.north, anchor=west, rectangle split,rectangle split parts=6, draw=latentcolor!50!black, inner sep=0pt, outer sep=0pt,rectangle split part fill={latentcolor}] {};
    \path (enc) edge[-latex] (latent);
    \node (dec) [right=3mm of latent.east, anchor=north, trapezium, trapezium angle=65, minimum width=20mm, minimum height=6mm, draw=autoencodercolor!50!black, thick, rotate=90, fill=autoencodercolor] {};
    \path (latent) edge[-latex] (dec);
    \node (dectext) [anchor=center]  at (dec.center)  {$W_{\text{dec}}$};
    \node (reconstructed) [posestyle,right=3mm of dec.south, anchor=west] {};
    \path (dec) edge[-latex] (reconstructed);
    \node (regloss) [above = 6.5mm of latent.north, losscolor] {$\mathcal{L}_{\text{reconstr}}$};
    \draw [-latex,losscolor,dashed] (poses.north) to [out=45,in=180] (regloss.west);
    \draw [-latex,losscolor,dashed] (reconstructed.north) to [out=135,in=0] (regloss.east);
    \node (sparseenc) [below = 13mm of enctext.center, anchor=center, losscolor] {$\mathcal{L}_{\text{sparse}}$};
    \draw [-latex,losscolor,dashed] (enctext.south) to (sparseenc.north);
    \node (sparsedec) [below = 13mm of dectext.center, anchor=center, losscolor] {$\mathcal{L}_{\text{sparse}}$};
    \draw [-latex,losscolor,dashed] (dectext.south) to (sparsedec.north);
\end{tikzpicture}\hspace*{11pt}}
\hfill
&
\subcaptionbox{Step 3: Fine-tune the model for consistency.\label{fig:tune_with_reg}}{
\begin{tikzpicture}[font=\small]
    \node (image) [rectangle, draw, minimum width=12mm, minimum height=10mm]{Image};
    \node (net) [right=3mm of image.east, anchor=south, trapezium, trapezium angle=80, minimum width=20mm, minimum height=8mm, draw=netcolor!50!black, thick, rotate=-90,fill=netcolor] {};
    \node (nettext) [anchor=center,align=center]  at (net.center)  {3D Pose \\ Estimator};
    \path (image) edge[-latex] (net);
    \node (poses) [posestyle,right=3mm of net.north, anchor=west] {};
    \path (net) edge [-latex] (poses);
    \node (loss) [below right= 13mm and 10mm of poses.center, anchor=center, losscolor]{$\mathcal{L}_{\text{pose}}$};
    \draw [-latex,losscolor,dashed] (poses.south) to [out=270,in=180] (loss.west);
    \node (enc) [right=3mm of poses.east, anchor=south, trapezium, trapezium angle=65, minimum width=20mm, minimum height=6mm, draw=autoencodercolor!50!black, thick, rotate=-90,pattern=north west lines, pattern color=autoencodercolor] {};
    \path (poses) edge[-latex] (enc);
    \node (enctext) [anchor=center,fill=white, rounded corners, inner sep=3pt]  at (enc.center)  {$W_{\text{enc}}$};
    \node (latent) [right=3mm of enc.north, anchor=west, rectangle split,rectangle split parts=6, draw=latentcolor!50!black, inner sep=0pt, outer sep=0pt,rectangle split part fill={latentcolor}] {};
    \path (enc) edge[-latex] (latent);
    \node (dec) [right=3mm of latent.east, anchor=north, trapezium, trapezium angle=65, minimum width=20mm, minimum height=6mm, draw=autoencodercolor!50!black, thick, rotate=90,pattern=north west lines, pattern color=autoencodercolor] {};
    \path (latent) edge[-latex] (dec);
    \node (dectext) [anchor=center,fill=white, rounded corners, inner sep=3pt]  at (dec.center)  {$W_{\text{dec}}$};
    \node (reconstructed) [posestyle,right=3mm of dec.south, anchor=west] {};
    \path (dec) edge[-latex] (reconstructed);
    \node (regloss) [above = 6.5mm of latent.north, losscolor] {$\mathcal{L}_{\text{cons}}$};
    \draw [-latex,losscolor,dashed] (poses.north) to [out=45,in=180] (regloss.west);
    \draw [-latex,losscolor,dashed] (reconstructed.north) to [out=135,in=0] (regloss.east);
    \node (frozen) [below = 8mm of latent.center, anchor=center, latentcolor] {\footnotesize  Frozen};
    \draw [-latex,latentcolor, thick] (frozen.west) to [out=180,in=-90] (enctext.south);
    \draw [-latex,latentcolor, thick] (frozen.east) to [out=0,in=-90] (dectext.south);
    \node (select) [anchor=center, draw=latentcolor, dashed, rounded corners, thick, minimum width=4mm, minimum height=22mm] at (poses.center){};
    \node (selecttext) [left = 15mm of regloss.west, anchor=east, latentcolor] {\footnotesize{Final Output}};
    \draw [-latex,latentcolor] (selecttext.east) to [out=0,in=90] (select);
    
\end{tikzpicture}}
\end{tabular}
\caption{Our complete training workflow. We train an initial model on all skeletons of multiple datasets without enforcing consistency. Using this model, we create pseudo-ground truth, needed to train an autoencoder that learns a latent keypoint space.
In turn, we use this frozen autoencoder to regularize the initial model during fine-tuning, encouraging consistent predictions.}
\label{fig:training_proceedure}
\end{figure*}

%% file: figures/latent_and_latentpred_model.tex
\tikzset{posestylesmall/.style={rectangle split,rectangle split parts=8, draw, inner sep=0pt, outer sep=0pt,rectangle split part fill={dsa,dsa,dsb,dsb,dsc,dsd,dse}}}
\begin{figure}
\subcaptionbox{Direct prediction of latent keypoints.\label{fig:latent_prediction_arch}}{
\begin{tikzpicture}[font=\small]
    \node (image) [rectangle, draw, minimum width=12mm, minimum height=10mm]{Image};
    \node (net) [right=3mm of image.east, anchor=south, trapezium, trapezium angle=80, minimum width=20mm, minimum height=8mm, draw=netcolor!50!black, thick, rotate=-90,fill=netcolor] {};
    \node (nettext) [anchor=center,align=center]  at (net.center)  {3D Pose \\ Estimator};
    \path (image) edge[-latex] (net);
    \node (latent) [right=15mm of net.north, anchor=center, rectangle split,rectangle split parts=4, draw=latentcolor!50!black, inner sep=0pt, outer sep=0pt,rectangle split part fill={latentcolor}] {};
    \path (net.north) edge[-latex](latent);
    \node (dec) [right=3mm of latent.east, anchor=north, trapezium, trapezium angle=65, minimum width=15mm, minimum height=6mm, draw=autoencodercolor!50!black, thick, rotate=90,pattern=north west lines, pattern color=autoencodercolor] {};
    \path (latent) edge[-latex] (dec);
    \node (reconstructed) [posestylesmall,right=3mm of dec.south, anchor=west] {};
    \path (dec) edge[-latex] (reconstructed);
    \node (loss) [right=4.8mm of reconstructed.center, anchor=west, losscolor]{$\mathcal{L}_{\text{pose}}$};
    \draw [-latex,losscolor,dashed] (reconstructed.east) to (loss.west);
\end{tikzpicture}}

\subcaptionbox{Hybrid of Fig.~\ref{fig:tune_with_reg} and \ref{fig:latent_prediction_arch} with a further student-teacher loss.\label{fig:student_teacher_arch}}{
\begin{tikzpicture}[font=\small]
    \node (image) [rectangle, draw, minimum width=12mm, minimum height=10mm]{Image};
    \node (net) [right=3mm of image.east, anchor=south, trapezium, trapezium angle=80, minimum width=20mm, minimum height=8mm, draw=netcolor!50!black, thick, rotate=-90,fill=netcolor] {};
    \node (nettext) [anchor=center,align=center]  at (net.center)  {3D Pose \\ Estimator};
    \path (image) edge[-latex] (net);
    \node (poses) [posestylesmall,above right=8mm and 6mm of net.north, anchor=west] {};
    \draw [-latex] (net.north) + (0,2.5mm) to [out=0, in=180] (poses);
    %\node (loss) [below right= 13mm and 10mm of poses.center, anchor=center, losscolor]{$\mathcal{L}_{pose}$};
    %\draw [-latex,losscolor,dashed] (poses.south) to [out=270,in=180] (loss.west);
    \node (enc) [right=3mm of poses.east, anchor=south, trapezium, trapezium angle=65, minimum width=15mm, minimum height=6mm, draw=autoencodercolor!50!black, thick, rotate=-90,pattern=north west lines, pattern color=autoencodercolor] {};
    \path (poses) edge[-latex] (enc);
    \node (latent) [right=3mm of enc.north, anchor=west, rectangle split,rectangle split parts=4, draw=latentcolor!50!black, inner sep=0pt, outer sep=0pt,rectangle split part fill={latentcolor}] {};
    \path (enc) edge[-latex] (latent);
    \node (dec) [right=3mm of latent.east, anchor=north, trapezium, trapezium angle=65, minimum width=15mm, minimum height=6mm, draw=autoencodercolor!50!black, thick, rotate=90,pattern=north west lines, pattern color=autoencodercolor] {};
    \path (latent) edge[-latex] (dec);
    \node (reconstructed) [posestylesmall,right=3mm of dec.south, anchor=west] {};
    \path (dec) edge[-latex] (reconstructed);
    %\node (loss2) [below left= 13mm and 10mm of reconstructed.center, anchor=center, losscolor]{$\mathcal{L}_{pose}$};
    %\draw [-latex,losscolor,dashed] (reconstructed.south) to [out=270,in=0] (loss2.east);
    \node (regloss) [above = 4.5mm of latent.north, losscolor] {$\mathcal{L}_{\text{cons}}$};
    \draw [-latex,losscolor,dashed] (poses.north) to [out=45,in=180] (regloss.west);
    \draw [-latex,losscolor,dashed] (reconstructed.north) to [out=135,in=0] (regloss.east);
    \node (latent2) [below=12mm of latent.south, anchor=north, rectangle split,rectangle split parts=4, draw=latentcolor!50!black, inner sep=0pt, outer sep=0pt,rectangle split part fill={latentcolor}] {};
    \draw [-latex] (net.north) + (0,-2.5mm) to [out=0, in=180] (latent2);
    \node (dec2) [right=3mm of latent2.east, anchor=north, trapezium, trapezium angle=65, minimum width=15mm, minimum height=6mm, draw=autoencodercolor!50!black, thick, rotate=90,pattern=north west lines, pattern color=autoencodercolor] {};
    \path (latent2) edge[-latex] (dec2);
    \node (reconstructed2) [posestylesmall,right=3mm of dec2.south, anchor=west] {};
    \path (dec2) edge[-latex] (reconstructed2);
    \node (loss) [left=25mm of regloss.center, anchor=center, losscolor]{$\mathcal{L}_{\text{pose}}$};
    \draw [-latex,losscolor,dashed] (poses.north) to [out=90,in=0] (loss.east);
    \node (loss2) [right=4.8mm of reconstructed.center, anchor=west, losscolor]{$\mathcal{L}_{\text{pose}}$};
    \draw [-latex,losscolor,dashed] (reconstructed.east) to (loss2.west);
    \node (loss3) [right=4.8mm of reconstructed2.center, anchor=west, losscolor]{$\mathcal{L}_{\text{pose}}$};
    \draw [-latex,losscolor,dashed] (reconstructed2.east) to (loss3.west);
    %\path (latent.south) edge[-latex,losscolor,dashed]  node [pos=0.45,right] {$\mathcal{L}_{teach}$} (latent2.north) {};
    \node (lossteach) [anchor=center,losscolor] at ($(latent)!0.5!(latent2)$) {$\mathcal{L}_{\text{teach}}$};
    \draw [-latex,losscolor,dashed] (latent.south) to (lossteach.north);
    \draw [-latex,losscolor,dashed] (latent2.north) to (lossteach.south);
    \node (select) [anchor=center, draw=latentcolor, dashed, rounded corners, thick, minimum width=4mm, minimum height=16mm] at (reconstructed2.center){};
    \node (selecttext) [above right=6mm and 3mm of select.east, anchor=west, latentcolor,align=left] {\footnotesize Final \\[-2pt]\footnotesize Output};
    \draw [-latex,latentcolor] (selecttext.west) -- ++(-3mm,0);
\end{tikzpicture}}
\caption{Alternative model structures for the fine-tuning phase, to be used instead of \reffig{tune_with_reg} in our training workflow.}
\label{fig:extra_archs}
\end{figure}

%% file: includes/tab_data_sizes.tex
\begin{table*}
\small
\newcommand{\metrics}{\tiny MPJPE$\downarrow$ & \tiny PMPJPE$\downarrow$ & \tiny PCK\textsubscript{100}$\uparrow$ & \tiny CPS\textsubscript{200}$\uparrow$}
\caption{
    Results using different amounts of datasets when training a separate-head model. \reftab{datasets} defines which datasets belong in which combination size.
    Using more datasets improves results on the 3DPW, 3DHP and MuPoTS benchmarks.
    On Human3.6M the small dataset combination gives better results, but this studio benchmark it less suited for studying real-world generalization capacity, as opposed to in-the-wild and outdoor benchmarks such as 3DPW and MuPoTS.
}
\label{tab:dataset_sizes}
\setlength{\tabcolsep}{2.0pt}
\renewcommand{\arraystretch}{0.92}
\begin{tabularx}{\textwidth}{l YYYY c YYYY c YYYY c YYYY}
    \toprule
    & \multicolumn{4}{c}{MuPoTS-3D}&& \multicolumn{4}{c}{3DPW} && \multicolumn{4}{c}{MPI-INF-3DHP} && \multicolumn{4}{c}{Human3.6M} \\
    \cmidrule{2-5} \cmidrule{7-10} \cmidrule{12-15} \cmidrule{17-20}
    & \metrics&& \metrics && \metrics && \metrics \\
    \midrule
    Single dataset & 91.3 & 62.9 & 65.3 & 53.5 && -- & -- & -- & -- && 66.5 & 46.7 & 83.0 & 78.4 && 48.3 & \f{33.2} & 92.1 & 89.8 \\
    \midrule
    Small (\sets) & 88.4 & 61.3 & 67.3 & 59.9 && 81.0 & 54.5 & 72.9 & 35.8 && 64.8 & 45.9 & 83.7 & 78.9 && \f{42.1} & 33.8 & \f{94.6} & \f{90.3} \\
    Medium (\setm) & 86.1 & 59.4 & 69.0 & \f{67.9} && 64.3 & 45.6 & 82.5 & 70.0 && 61.7 & 44.6 & 85.6 & 80.2 && 43.2 & 34.5 & 94.3 & 90.3 \\
    Full (\setl) & \f{84.6} & \f{59.0} & \f{70.1} & 66.0 && \f{61.8} & \f{43.4} & \f{83.8} & \f{71.1} && \f{59.6} & \f{44.1} & \f{86.6} & \f{81.8} && 44.7 & 34.3 & 94.3 & 90.1 \\
    \bottomrule
\end{tabularx}
\end{table*}

%% file: includes/tab_main.tex
\begin{table*}
\small
\newcommand{\metrics}{\tiny MPJPE$\downarrow$ & \tiny PMPJPE$\downarrow$ & \tiny PCK\textsubscript{100}$\uparrow$ & \tiny CPS\textsubscript{200}$\uparrow$}
\caption{Evaluation of different strategies for handling different skeleton annotation formats during training.}
\label{tab:approach_comparisons}
\setlength{\tabcolsep}{1.5pt}
\renewcommand{\arraystretch}{0.92}
\begin{tabularx}{\textwidth}{cl YYYY c YYYY c YYYY c YYYY}
\toprule
&& \multicolumn{4}{c}{MuPoTS-3D}&& \multicolumn{4}{c}{3DPW} && \multicolumn{4}{c}{MPI-INF-3DHP} && \multicolumn{4}{c}{Human3.6M} \\
\cmidrule{3-6} \cmidrule{8-11} \cmidrule{13-16} \cmidrule{18-21}
&& \metrics&& \metrics && \metrics && \metrics \\
\midrule
\multirow{5}{8pt}{\rotatebox{90}{\scriptsize EffNetV2-S~~~~~}}
&Merged joints & 91.9 & 67.3 & 63.2 & 69.9 && 72.5 & 48.3 & 79.5 & 69.7 && 69.8 & 51.6 & 80.7 & 79.4 && \f{44.6} & 34.2 & 93.9 & 89.8 \\
&Separate joints (F.~\ref{fig:train_init}) & 84.6 & 59.0 & 70.1 & 66.0 && 61.8 & 43.4 & 83.8 & 71.1 && 59.6 & 44.1 & \f{86.6} & 81.8 && 44.7 & 34.3 & 94.3 & \f{90.1} \\
\cmidrule{2-21}
&Consistency regul. (F.~\ref{fig:tune_with_reg}) & \f{81.8} & \f{57.8} & \f{72.5} & \f{72.9} && \f{61.5} & \f{43.0} & \f{84.0} & \f{71.9} && \f{59.2} & \f{43.6} & \f{86.6} & \f{82.7} && 45.2 & \f{33.3} & \f{94.4} & \f{90.1} \\
&Latent pred. (F.~\ref{fig:latent_prediction_arch}) & 83.0 & 58.9 & 71.4 & 71.2 &&  62.0 & 43.6 & \f{84.0} & 71.7 && 60.2 & 44.7 & 86.1 & 80.2 && 46.5 & 34.4 & 93.9 & 89.5 \\
&Hybrid (F.~\ref{fig:student_teacher_arch}) & 82.7 & 58.5 & 71.6 & 72.1 && 61.8 &  43.3 & \f{84.0} & 71.8 && 60.4 & 44.8 & 85.9 & 80.9 && 46.1 & 34.2 & 94.1 & 89.4 \\
\midrule
\multirow{3}{8pt}{\rotatebox{90}{\scriptsize EffNetV2-L}}
&Separate joints & 82.9 & 57.7 & 71.0 & 70.9 &&  60.9 & 42.1 & 84.4 & 73.4 && 59.1 & 42.2 & 88.0 & \f{85.3} && 41.6 & 32.0 & 95.1 & 92.1 \\
\cmidrule{2-21}
&Consistency regul. & \f{81.0} & \f{57.4} & \f{72.8} & \f{74.8} && \f{60.6} & \f{41.7} & \f{84.7} & \f{74.3} && \f{57.9} & \f{41.8} & \f{88.2} & 84.7 && \f{40.6} & \f{30.7} & \f{95.7} & \f{92.6} \\
&Hybrid & 81.3 & 57.9 & 72.4 & 73.9 &&  61.1 & 42.0 & 84.6 & \f{74.3} && 59.2 & 42.8 & 87.2 & 84.3 && 41.8 & 31.4 & 95.6 & \f{92.6} \\
\bottomrule
\end{tabularx}
\end{table*}

%% file: includes/tab_sota.tex
\begin{table}
\caption{Comparison to recent state-of-the-art works. (*twice as long training, 384 px resolution)}
\label{tab:sota}
\setlength{\tabcolsep}{3pt}
\renewcommand{\arraystretch}{0.92}
\footnotesize
\begin{tabularx}{\linewidth}{l c p{3pt} YYY p{3pt} YY p{3pt} c}
    \toprule
    & \tiny MuPoTS && \multicolumn{3}{c}{\tiny 3DPW} &&\multicolumn{2}{c}{\tiny 3DHP} && \tiny H3.6M \\
    \cmidrule{2-2} \cmidrule{4-6}  \cmidrule{8-9}  \cmidrule{11-11}
    & \tiny PCK\textsubscript{150}$\uparrow$ && \tiny MPJPE$\downarrow$ & \tiny PMPJPE$\downarrow$ & \tiny PCK\textsubscript{50}$\uparrow$ && \tiny MPJPE$\downarrow$ & \tiny PCK\textsubscript{150}$\uparrow$ &&  \tiny MPJPE$\downarrow$ \\
    \midrule
    ROMP~\cite{Sun21ICCV}      & --   && 80.1 & 56.8 & 36.5 && --   & --   && --   \\
    Lin~\cite{Lin21ICCV}       & --   && 74.7 & 45.6 & --   && --   & --   && 51.2 \\
    PoseAug~\cite{Gong21CVPR}  & --   && --   & --   & --   && 71.1 & 89.2 && 50.2 \\
    Cheng~\cite{Cheng22PAMI}   & 89.6 && --   & --   & --   && --   & --   && 49.3 \\
    \midrule
    Ours RN50                  & 92.2 && 65.5 & 47.2 & 49.0 && 64.2 & 93.3 && 45.8 \\
    Ours EffV2S                & 93.7 && 61.5 & 43.0 & 51.8 && 60.0 & 95.3 && 45.2 \\
    Ours EffV2L                & 94.1 && 60.6 & 41.7 & 52.1 && 59.2 & 95.8 && 40.6 \\    
    Ours EffV2L*               & 95.4 && 58.9 & 39.5 & 53.9 && 55.4 & 97.1 && 36.5 \\    
    \bottomrule
\end{tabularx}
\end{table}

%% file: includes/tab_chirality_effect.tex
\begin{table}
\newcommand{\metrics}{\tiny MPJPE$\downarrow$ & \tiny PCK\textsubscript{100} & \tiny CPS\textsubscript{200}}
\caption{Effect of enforcing chirality equivariance constraints on the autoencoder weight matrices.}
\label{tab:chirality_effect}
\setlength{\tabcolsep}{2.0pt}
\renewcommand{\arraystretch}{0.92}

\footnotesize
\begin{tabularx}{\linewidth}{l c YYY c YYY c YYY}
\toprule
&\multirow{2}{10pt}{\hspace*{3pt}\rotatebox{90}{\tiny Chirality~}} & \multicolumn{3}{c}{MuPoTS} && \multicolumn{3}{c}{3DPW} && \multicolumn{3}{c}{3DHP} \\
\cmidrule{3-5} \cmidrule{7-9} \cmidrule{11-13}
&& \metrics && \metrics  && \metrics\\
\midrule
Cons. regul.     &              & \f{81.8} &    72.4  & \f{73.1} &&    61.6  &    83.9  &    71.9  && \f{59.2} & \f{86.6} &    82.1  \\
Cons. regul.     & $\checkmark$ & \f{81.8} & \f{72.5} &    72.9  && \f{61.5} & \f{84.0} &    71.9  && \f{59.2} & \f{86.6} & \f{82.7} \\
Hybrid           &              &    83.2  &    71.2  &    71.7  &&    61.7  & \f{84.0} & \f{72.0} &&    60.3  &    85.9  &    80.7  \\
Hybrid           & $\checkmark$ &    82.7  &    71.6  &    72.1  &&    61.8  & \f{84.0} &    71.8  &&    60.4  &    85.9  &    80.9  \\
\bottomrule
\end{tabularx}
\end{table}

%% file: includes/tab_latent_size.tex
\begin{table}
\newcommand{\metrics}{\tiny MPJPE$\downarrow$ & \tiny PCK\textsubscript{100}$\uparrow$ & \tiny CPS\textsubscript{200}$\uparrow$}
\caption{Effect of the number of latent points on final performance.
We use 48 as our default setting.}
\label{tab:latent_size}
\setlength{\tabcolsep}{2.0pt}
\renewcommand{\arraystretch}{0.93}
\footnotesize
\begin{tabularx}{\linewidth}{cl YYY c YYY c YYY}
\toprule
& \multirow{2}{10pt}{\hspace*{3pt}\rotatebox{90}{\tiny \#latents~~}}& \multicolumn{3}{c}{MuPoTS} && \multicolumn{3}{c}{3DPW} && \multicolumn{3}{c}{3DHP} \\
\cmidrule{3-5} \cmidrule{7-9} \cmidrule{11-13}
&& \metrics && \metrics  && \metrics\\
\midrule
\multirow{4}{7pt}{\rotatebox{90}{\scriptsize Cons. regul.}}
& 24 & \f{81.6} &    72.4  & \f{73.1} &&    62.0  &    83.9  &    71.7  && \f{58.9} &    86.5  &    81.7  \\
& 32 &    82.2  &    72.1  &    72.5  &&    61.8  &    83.9  &    71.8  &&    59.2  &    86.5  &    81.9  \\
& 48 &    81.8  & \f{72.5} &    72.9  && \f{61.5} & \f{84.0} & \f{71.9} &&    59.2  & \f{86.6} & \f{82.7} \\
& 64 &    82.3  &    72.0  &    73.0  &&    61.8  &    83.8  &    71.8  &&    59.2  & \f{86.6} &    82.1  \\
\midrule
\multirow{4}{7pt}{\rotatebox{90}{\scriptsize Hybrid}}
& 24 &    86.0  &    69.4  &    65.0  &&    67.2  &    81.2  &    64.6  &&    70.2  &    80.0  &    62.2  \\
& 32 & \f{82.7} &    71.4  & \f{72.1} &&    62.3  &    83.9  & \f{71.8} && \f{60.2} & \f{86.2} & \f{81.4} \\
& 48 & \f{82.7} & \f{71.6} & \f{72.1} && \f{61.8} & \f{84.0} & \f{71.8} &&    60.4  &    85.9  &    80.9  \\
& 64 &    84.1  &    70.6  &    67.0  &&    62.2  &    83.7  &    71.3  &&    62.0  &    84.5  &    80.2  \\
\bottomrule
\end{tabularx}
\end{table}

%% file: supp_body.tex
\begin{abstract}
In the supplementary material, we provide additional qualitative results, as well as details about implementation, data processing and evaluation. We also provide two additional ablations (training length and batch norm configuration) and a derivation for the claim that $\ell_1$ regularization of the affine-combining autoencoder leads to a reduction in negative weights and hence near-convex combinations. 
\end{abstract}

\section{Additional Qualitative Results}
Similar to the qualitative results shown in the main paper, Figures~\ref{fig:qualitative_results1}, \ref{fig:qualitative_results2}, and \ref{fig:qualitative_results3} show further predictions for a variety of images.
It can clearly be seen that the model with separate heads without consistency regularization creates rather inconsistent skeleton predictions, whereas fine-tuning with our ACAE regularization significantly improves the consistency.
Furthermore, these figures show that the resulting models display excellent in-the-wild performance, even on challenging poses, or in suboptimal lighting conditions.

\section{Training Details}

\begin{figure}[t]
\centering
\includegraphics[width=\linewidth]{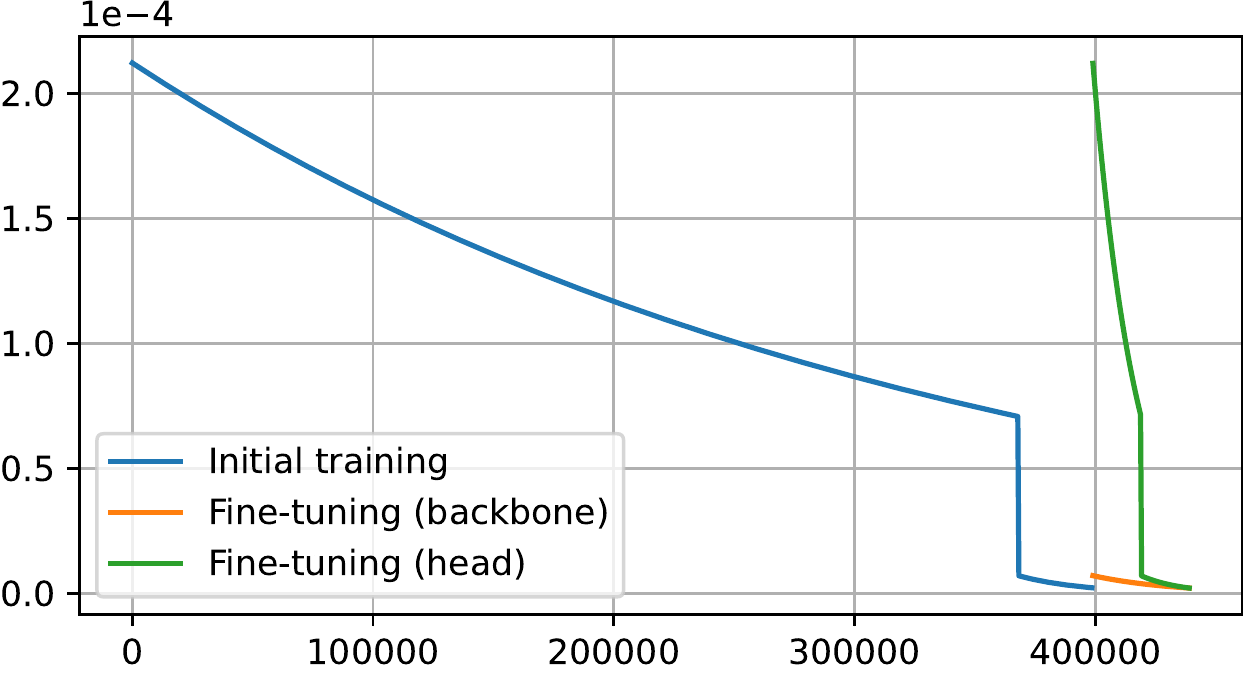}
\caption{Learning rate schedule.}
\label{fig:lr_sched}
\end{figure}

\PAR{Learning Rate.} Our learning rate schedule is shown in \reffig{lr_sched}.
The learning rate starts at 2.12e-4 and exponentially decays by a factor of 3 over 92\% of training, then drops by a factor of 10 and then further decays exponentially by a factor of 3 until the end of the initial training.

The fine-tuning phase uses two different learning rates.
We perform a warm restart on the last layer (the prediction head) in order to ensure that the regularization loss can take effect, without disrupting the already mostly converged weights of the backbone.
For the head, we follow a similar recipe as in the initial training, but we perform the large learning rate drop at 50\% of the fine-tuning phase.
For the backbone, we repeat the last, decaying segment of the initial schedule.

\PAR{Loss Details.} 
We make minor adjustments to the MeTRAbs model~\cite{Sarandi21TBIOM}, which we use as the basis of our experiments.
In \cite{Sarandi21TBIOM}, the authors perform internal supervision on the output of a 2D heatmap head and a 3D heatmap head.
Instead, we simplify this and only use the absolute pose output for supervision, \ie, the 2D projection loss and the mean-relative loss are computed on this single, absolute output.
This makes the implementation cleaner when more losses are added for consistency regularization or student-teacher latent matching, etc., since the model can be treated as a single-output black box.

The weak supervision loss for 2D-annotated examples only consists of the 2D projection loss.
For this, we do not specifically predict skeletons according to the skeleton formats of the 2D datasets.
Instead, the prediction is derived by averaging the corresponding 3D joint predictions for every output skeleton format. In other words, for the calculation of the 2D weak loss, we consider our prediction for the left shoulder to be the average of all the left shoulder joints in every skeleton format that we use.

We use $\lambda_{\text{proj}}=1$ and $\lambda_{\text{abs}}=0.1$ and scale the weak-supervision-loss by a factor of 0.2.

The absolute loss $\mathcal{L}_{\text{abs}}$ is only turned on after 5000 steps (also in fine-tuning, for consistency), similarly with the teacher loss.
In some datasets the absolute distance to the person can be very large (\eg, JTA, SAIL-VOS, ASPset).
Here the absolute loss would overwhelm the total loss, so we scale down the absolute Z component to a maximum effective distance of 10 m for loss computation.

\PAR{Batch Composition.} In \reftab{batchcomp}, we specify the number of examples from each dataset per batch.
This is based on the total number of examples in each dataset but not linearly, as we oversample smaller datasets compared to their size, in order to provide more diverse supervision to the model.
For batch generation, we set up one queue per dataset that iterates over epochs of that dataset, then we interleave the streams and chunk it into batches (as opposed to independently sampling each batch). 
\input{includes/tab_batch_composition_supp}

\PAR{Initialization Details.} We initialize with ImageNet-pretrained weights. For the RN50 experiment in Table 4 (SOTA), we use ResNet50V1.5 as implemented in PyTorch, ported to TensorFlow, along with the ImageNet weights, which we found to be superior to the ones provided with TensorFlow.

We precisely control the random seeds, which guarantees that bitwise equal batches are fed to each training run, improving comparability.

\PAR{Implementation Details.} We use TensorFlow 2.9 with Keras, CUDA 11.4 and CuDNN 8.2.4 for the implementation.
Training takes about 2 days with the EffV2-S backbone and about 6 days with EffV2-L on a single Nvidia A40 GPU (48 GB) in mixed FP16/FP32 precision.

\section{Data Processing Details}
Where missing, we obtain person bounding boxes with YOLOv4~\cite{yolov4:Bochkovskiy20Arxiv} and person segmentation with DeepLabv3~\cite{deeplabv3:Chen17Arxiv}.
Examples with implausible bone lengths are removed to avoid training on erroneous annotations.
We use all cameras of 3DHP, and all HD cameras of CMU-Panoptic (and all sequences with labels).
We further calibrated all cameras of BML-MoVi that did not have calibration provided in the dataset, and use all of them in training (based on pose predictions from an earlier version of our model).
We use 200k composited images for MuCo-3DHP, generated with the official Matlab script.

\section{Evaluation Details}
We evaluate all 24 SMPL joints for 3DPW, and all 17 joints for 3DHP and MuPoTS.
In case of 3DPW, the entire dataset is used for testing.
For 3DHP we use the official split, for H36M the most common split from the literature, \ie subjects S9 and S11 are used for testing.

For MuPoTS, we evaluate the matched poses.
As we use the same YOLOv4 detector in all our experiments, we have 94.6\% recall in all of our experiments (hence the matched-pose results are directly comparable).
For our main evaluations, in each benchmark, we simply calculate the average metrics over all metric-scale poses.

In Table 4 (SOTA comparison) of the main paper, we use the more complex standard evaluation metrics.
That is, for MuPoTS, here we use bone rescaling, normalized skeletons, and averaging is performed first per sequence and the final value is the average of per-sequence averages.
In this, and also other details, we follow the same protocols as \cite{Sarandi21TBIOM} (\eg, which joints to evaluate). 

% 3DPW: 68663
% H36M: 8724
% MuPoTS: 19633
% 3DHP: 2875
% 17 joints

\section{Additional Ablations}
\PAR{Training Length.} In \reftab{trainlen}, we study the effect of the length of training on the final model performance.
Clearly, longer training can further improve the results and especially the correct pose score improves.
Do note that every new line doubles the number of training steps, so this is expensive.
Further, with long trainings we noted that training in 16-bit floating point (FP16) precision is unstable, as the activations tend to grow out of the representable range.
The cause of this was that the convolutional kernels of the backbone grow in scale during gradient descent, since.
Since they are always followed by BatchNorm in EfficientNetV2, the weight scale has no impact on the network output (though it has an effect on the effective learning rate~\cite{VanLaarhoven17Arxiv}).
We mitigated the problem by applying a max-norm constraint on the convolutional kernels to keep them from growing without bound.
Some numerical instability remains in case of very long trainings, the cause of which needs more investigation.
\reftab{finetunelen} shows that further extending the fine-tuning phase can bring minor performance benefits.
For reasons of practicality, we chose 400k training steps and 40k fine-tuning step as the default setting for all of our experiments in the main paper, albeit one could achieve slightly better results with longer schedules.
\input{includes/tab_trainlen_supp}

\PAR{Ghost BatchNorm.}
In \reftab{ghostbn} we show an ablation on using Ghost BatchNorm~\cite{Hoffer17NIPS,Summers19ICLR}.
We compare three options: normal BatchNorm, Ghost BN where the 96 3D annotated examples are normalized as one group and the 32 2D-labeled ones as another, and Ghost BN with ghost batch size 16.
While the differences are not very large, the Ghost BN options tend to perform better.
This is probably due to the discrepancies in BatchNorm statistics among datasets.

\PAR{Inference-Mode BatchNorm Fine-Tuning.}
Furthermore, \reftab{ghostbn} also demonstrates that, when using Ghost BN, it is important to fine tune the network at the end in inference mode.
By inference-mode fine-tuning, we mean that the BN layers use the stored, fixed statistics for normalization instead of the usual training mode of using the statistics of the current minibatch.
In Ghost BN, the stored statistics may be suboptimal, since they are updated based on parts of the batch, instead of the overall batch statistics.
A final fine-tuning in ``inference mode'' allows the network to fine-tune its weights to the setting that it will be used in during inference (\ie, to adapt the weights to work well with the stored statistics).
\input{includes/tab_batchnorms_supp}

\section{Effects of L1 Regularization in the ACAE}
We point out in the main paper that using $\ell_1$ regularization on the weight matrices of the affine-encoding autoencoder results both in sparsity and fewer negative weights.
Here we elaborate on this connection.
Since the weights produce affine combinations, they sum to one as specified in Eq. 1-2 in the main paper.

\begin{align}
\sum_{j=1}^{J} w_{l,j}^{\text{enc}} &= 1 \\
\sum_{l=1}^{L} w_{j,l}^{\text{dec}} &= 1.
\end{align}

We can partition the weights to negative and non-negative ones.

\begin{align}
w_{l, +}^{\text{enc}} &= \sum_{j \ : \ w_{l,j}^{\text{enc}}  \geq 0} w_{l,j}^{\text{enc}} \\
w_{l, -}^{\text{enc}} &= \sum_{j\ : \ w_{l,j}^{\text{enc}}  < 0} w_{l,j}^{\text{enc}} \\
w_{l, +}^{\text{enc}} + w_{l, -}^{\text{enc}} &=1,
\end{align}
and analogously for the decoder weights.
Now, the $\ell_1$ penalty (sum of absolute values) can be written as

\begin{align}
\ell_1(w_{l,\cdot}^{\text{enc}}) &= \sum_{j=1}^{J} \left| w_{l,j}^{\text{enc}} \right|  = \\
 &= w_{l, +}^{\text{enc}} - w_{l, -}^{\text{enc}} = \\
 &= \left(1-w_{l, -}^{\text{enc}}\right) - w_{l, -}^{\text{enc}} = \\
 &= 1-2 \cdot w_{l, -}^{\text{enc}} = \\
  &= 1+2 \cdot \left| w_{l, -}^{\text{enc}} \right|.
\end{align}

This means that the $\ell_1$ penalty is equivalent to penalizing the absolute sum of the negative weights.

When all weights are non-negative, we get convex combinations.
In other words, the $\ell_1$ regularization in the ACAE encourages constructing close-to-convex combinations besides sparsity.

\include{includes/fig_qualitative_results_supp}

%% file: includes/tab_batch_composition_supp.tex
\begin{table}[t]
\caption{
    Batch composition for the experiments with the three different levels of dataset combinations. Each minibatch consists of 96 examples with 3D labels and 32 with 2D labels.
}
\setlength\tabcolsep{4mm}
\centering
\begin{tabularx}{\linewidth}{lccc}
\toprule
Dataset name & Small & Medium & Full  \\
\midrule
\multicolumn{4}{c}{\textit{Real images with markerless MoCap}} \vspace{0.9mm}\\
\gb MuCo-3DHP & 32 & 9 & 6 \\
CMU-Panoptic  & -- & 9 & 7 \\
\gb AIST-Dance++ & -- & 9 & 6 \\
HUMBI & -- & 7 & 5 \\
\gb MPI-INF-3DHP & -- & 5 & 3 \\
RICH  & -- & 7 & 4 \\
\gb BEHAVE & -- & -- & 3 \\
ASPset & -- & -- & 4 \\
\gb 3DOH50K & -- & -- & 3 \\
IKEA ASM & -- & -- & 2 \\
\midrule
\multicolumn{4}{c}{\textit{Real images with marker-based MoCap}}\vspace{0.9mm}\\
\gb Human3.6M & 32 & 9 & 4 \\
TotalCapture & -- & 5 & 3 \\
\gb BML-MoVi & -- & -- & 5 \\
Berkeley-MHAD & -- & -- & 3 \\
\gb UMPM & -- & --  & 2 \\
Fit3D & -- & -- & 2 \\
\gb GPA & -- & -- & 4 \\
HumanSC3D & -- & -- & 1 \\
\gb CHI3D & -- & -- & 1 \\
Human4D & -- & -- & 1 \\
\gb MADS & -- & -- & 2 \\
\midrule
\multicolumn{4}{c}{\textit{Synthetic images}}\vspace{0.9mm}\\
\gb SURREAL & 32 & 8 & 5 \\
3DPeople & -- & 6 & 4 \\
\gb JTA & -- & 5 & 3 \\
HSPACE & -- & 5 & 3 \\
\gb SAIL-VOS & -- & 7 & 5 \\
AGORA & -- & 5 & 3 \\
\gb SPEC & -- & -- & 2 \\
\midrule
\multicolumn{4}{c}{\textit{Real images with 2D annotations (weak supervision)}}\vspace{0.9mm}\\
\gb COCO & 8 & 8  & 8  \\
MPII & 8  & 8  & 8 \\
\gb PoseTrack & 8  & 8  & 8 \\
JRDB & 8  & 8  & 8 \\
\bottomrule
\end{tabularx}

\label{tab:batchcomp}
\end{table}

%% file: includes/tab_trainlen_supp.tex
\begin{table*}
\small
\newcommand{\metrics}{\tiny MPJPE$\downarrow$ & \tiny PMPJPE$\downarrow$ & \tiny PCK\textsubscript{100}$\uparrow$ & \tiny CPS\textsubscript{200}$\uparrow$}
\caption{
    Ablation for the length of training.
}
\label{tab:trainlen}
\setlength{\tabcolsep}{2.0pt}
\begin{tabularx}{\textwidth}{l YYYY p{5pt} YYYY p{5pt} YYYY p{5pt} YYYY}
    \toprule
    & \multicolumn{4}{c}{MuPoTS-3D}&& \multicolumn{4}{c}{3DPW} && \multicolumn{4}{c}{MPI-INF-3DHP} && \multicolumn{4}{c}{Human3.6M} \\
    \cmidrule{2-5} \cmidrule{7-10} \cmidrule{12-15} \cmidrule{17-20}
    & \metrics&& \metrics && \metrics && \metrics \\
    \midrule
    \multicolumn{20}{c}{\textit{Initial, separate-skeleton model}}\\
    100k           &    88.6   &    62.4   &    67.8   &    60.0   &&    65.9   &    47.0   &    81.5   &    65.3   &&    66.3   &    51.2   &    82.2   &    71.7   &&    47.7   &    37.6   &    91.9   &    86.6   \\
    200k           &    87.3   &    60.3   &    68.3   &    65.1   &&    63.1   &    44.6   &    82.6   &    69.5   &&    63.1   &    47.4   &    84.4   &    77.1   &&    46.8   &    36.2   &    93.1   &    88.7   \\
    400k (default) &    84.6   &    59.0   &    70.1   &    66.0   &&    61.8   &    43.4   & \f{83.8}  &    71.1   &&    59.6   &    44.1   &    86.6   &    81.8   &&    44.7   &    34.3   &    94.3   &    90.1   \\
    800k           & 82.9  & 57.8  & 70.5  & 69.8  && 61.7  & 42.6  &    83.7   & 73.1  && 58.8  & 42.9  & 87.3  & 83.0  && 43.0  & 33.2  & 94.8  & 91.4  \\
    1.6M            & \f{81.6}  & \f{56.7}  & \f{71.6}  & \f{72.8}  && \f{61.5}  & \f{41.9}  & \f{84.4}  & \f{74.2}  && \f{58.6}  & \f{41.3}  & \f{87.8}  & \f{86.3}  && \f{41.5}  & \f{32.3}  & \f{95.5}  & \f{92.1} \\
    \midrule
    \multicolumn{20}{c}{\textit{Fine-tuned with consistency regularization for 40k steps}}\\
    100k           &    85.5   &    60.6   &    70.3   &    66.6   &&    65.0   &    46.0   &    81.8   &    66.9   &&    63.6   &    48.7   &    83.9   &    74.1   &&    46.7   &    36.4   &    92.5   &    87.6   \\
    200k           &    84.2   &    59.2   &    70.8   &    70.4   &&    63.4   &    44.3   &    82.7   &    69.9   &&    61.4   &    46.3   &    85.4   &    79.1   &&    46.7   &    35.1   &    93.3   &    88.6   \\
    400k (default) &    81.8   &    57.8   &    72.5   &    72.9   &&    61.5   &    43.0   &    84.0   &    71.9   &&    59.2   &    43.6   &    86.6   &    82.7   &&    45.2   &    33.3   &    94.4   &    90.1   \\
    800k           & 80.5  & 56.8  & 72.7  & 74.4  && 61.3  & 42.1  & 84.5  & 73.3  && \f{57.7}  & 42.2  & 87.7  & 84.3  && 42.0  & 31.9  & 95.3  & 91.4  \\
    1.6M           & \f{79.7}  & \f{56.1}  & \f{73.3}  & \f{76.6}  && \f{60.6}  & \f{41.7}  & \f{84.7}  & \f{74.5}  && 58.6  & \f{41.1}  & \f{87.8}  & \f{86.6}  && \f{41.1}  & \f{31.2}  & \f{95.7}  & \f{92.2} \\
    \bottomrule
\end{tabularx}
\end{table*}

\begin{table*}
\small
\newcommand{\metrics}{\tiny MPJPE$\downarrow$ & \tiny PMPJPE$\downarrow$ & \tiny PCK\textsubscript{100}$\uparrow$ & \tiny CPS\textsubscript{200}$\uparrow$}
\caption{
    %EfficientNetV2-S trained with separate joint prediction.
    Ablation for the length of consistency-regularized fine-tuning with an initial training length of 400k steps.
}
\label{tab:finetunelen}
% Description & dataset 1 x3 metrics & dataset 2 x3 metrics & etc.
\setlength{\tabcolsep}{2.0pt}
\begin{tabularx}{\textwidth}{l YYYY p{5pt} YYYY p{5pt} YYYY p{5pt} YYYY}
    \toprule
    & \multicolumn{4}{c}{MuPoTS-3D}&& \multicolumn{4}{c}{3DPW} && \multicolumn{4}{c}{MPI-INF-3DHP} && \multicolumn{4}{c}{Human3.6M} \\
    \cmidrule{2-5} \cmidrule{7-10} \cmidrule{12-15} \cmidrule{17-20}
    & \metrics&& \metrics && \metrics && \metrics \\
    \midrule
    20k           &    82.0  &    58.0  &    72.2  &    72.1  &&    61.7  &    43.0  &    83.9  &    71.3  &&    59.3  &    43.6  &    86.5  &    82.1  &&    44.9  &    33.5  &    94.3  &    89.8  \\
    40k (default) &    81.8  &    \f{57.8}  & \f{72.5} &    72.9  &&    61.5  &    43.0  &    84.0  &    71.9  &&    59.2  &    43.6  &    86.6  &    82.7  &&    45.2  & \f{33.3} &    94.4  & \f{90.1} \\
    80k           & \f{81.6} & \f{57.8} &    72.4  & \f{73.2} && \f{61.4} & \f{42.9} & \f{84.1} & \f{72.1} && \f{58.4} & \f{43.2} & \f{87.2} & \f{82.9} && \f{44.5} & \f{33.3} & \f{94.6} & \f{90.1} \\
    \bottomrule
\end{tabularx}
\end{table*}

%% file: includes/tab_batchnorms_supp.tex
\begin{table*}
\small
\newcommand{\metrics}{\tiny MPJPE$\downarrow$ & \tiny PMPJPE$\downarrow$ & \tiny PCK\textsubscript{100}$\uparrow$ & \tiny CPS\textsubscript{200}$\uparrow$}
\caption{
    Ablation for Ghost Batch Normalization and inference-mode fine-tuning for 1000 steps.
}
\label{tab:ghostbn}
\setlength{\tabcolsep}{2.0pt}
\begin{tabularx}{\textwidth}{l YYYY p{5pt} YYYY p{5pt} YYYY p{5pt} YYYY}
    \toprule
    & \multicolumn{4}{c}{MuPoTS-3D}&& \multicolumn{4}{c}{3DPW} && \multicolumn{4}{c}{MPI-INF-3DHP} && \multicolumn{4}{c}{Human3.6M} \\
    \cmidrule{2-5} \cmidrule{7-10} \cmidrule{12-15} \cmidrule{17-20}
    & \metrics&& \metrics && \metrics && \metrics \\
    \midrule
    \multicolumn{20}{c}{\textit{Initial, separate-skeleton model, \textbf{with}  fine-tuning at the end with BN in inference mode}}\\
    Normal BN        & 84.5 & 59.2 & 70.0 & 65.9 && 62.6 & 43.6 & 83.6 & \bf71.9 && 61.3 & \bf43.7 & 85.7 & \bf82.0 && 46.3 & 34.6 & 94.2 & 89.9  \\
    Ghost BN (3D/2D) & \bf83.6 & \bf58.7 & \bf70.4 & \bf70.0 && 62.8 & 43.5 & 83.2 & 71.2 && 61.8 & 45.1 & 85.7 & 80.3 && 46.6 & 34.7 & 94.1 & \bf90.5  \\
    Ghost BN 16      & 84.6 & 59.0 & 70.1 & 66.0 && \bf61.8 & \bf43.4 & \bf83.8 & 71.1 && \bf59.6 & 44.1 & \bf86.6 & 81.8 && \bf44.7 & \bf34.3 & \bf94.3 & 90.1  \\
    \midrule
    \multicolumn{20}{c}{\textit{Initial, separate-skeleton model, \textbf{without} fine-tuning at the end with BN in inference mode}}\\
    Normal BN        & 84.2 & 59.1 & 70.4 & 66.4 && 62.6 & 43.6 & 83.5 & 71.9 && 60.4 & 43.4 & 86.1 & 82.3 && 45.9 & 34.4 & 94.2 & 89.8  \\
    Ghost BN (3D/2D) & 88.2 & 63.5 & 66.7 & 62.0 && 67.3 & 47.5 & 80.9 & 68.0 && 66.3 & 49.0 & 81.7 & 77.4 && 50.8 & 40.4 & 90.7 & 87.7  \\
    Ghost BN 16      & 85.8 & 60.4 & 69.0 & 63.5 && 63.4 & 44.7 & 83.2 & 70.8 && 59.8 & 44.3 & 86.3 & 81.8 && 45.4 & 35.7 & 93.6 & 89.6  \\
    \midrule
    \midrule
    \multicolumn{20}{c}{\textit{Fine-tuned with consistency regularization for 40k steps, \textbf{with}  fine-tuning at the end with BN in inference mode}}\\
    Normal BN        & 83.3 & 58.5 & 70.9 & 73.2 && 63.1 & 43.7 & 83.4 & 71.8 && 60.5 & \bf43.4 & 85.9 & \bf82.7 && 46.0 & 33.5 & \bf94.4 & 89.9 \\
    Ghost BN (3D/2D) & \bf81.2 & \bf57.7 & \bf72.5 & \bf74.0 && 62.2 & \bf43.0 & 83.7 & \bf72.3 && 60.5 & 44.4 & 86.1 & 80.9 && 46.2 & 33.7 & 94.1 & \bf90.6 \\
    Ghost BN 16      & 81.8 & 57.8 & \bf72.5 & 72.9 && \bf61.5 & \bf43.0 & \bf84.0 & 71.9 && \bf59.2 & 43.6 & \bf86.6 & \bf82.7 && \bf45.2 & \bf33.3 & \bf94.4 & 90.1 \\
    \midrule
    \multicolumn{20}{c}{\textit{Fine-tuned with consistency regularization for 40k steps, \textbf{without} fine-tuning at the end with BN in inference mode}}\\
    Normal BN        & 83.1 & 58.7 & 71.3 & 72.9 && 62.8 & 43.5 & 83.5 & 71.7 && 59.6 & 43.2 & 86.4 & 82.9 && 45.5 & 33.4 & 94.5 & 89.9 \\
    Ghost BN (3D/2D) & 89.1 & 64.6 & 66.4 & 62.1 && 68.5 & 49.4 & 79.6 & 65.2 && 73.4 & 52.9 & 76.4 & 70.9 && 58.4 & 43.4 & 85.7 & 82.7 \\
    Ghost BN 16      & 84.0 & 60.0 & 70.7 & 69.7 && 63.3 & 45.3 & 82.7 & 69.4 && 63.2 & 45.7 & 83.8 & 80.1 && 49.0 & 36.6 & 92.2 & 88.3 \\
    \bottomrule
\end{tabularx}
\end{table*}

%% file: includes/fig_qualitative_results_supp.tex
\newcommand{\cropim}[1]{\includegraphics[width=0.194\textwidth,trim={1.9cm 1.9cm 1.6cm 1.7cm},clip]{figures/qual_rep_seg2/pexels-#1.pdf}}
\newcommand{\figline}[1]{%
    \includegraphics[width=0.189\textwidth]{figures/qual_rep_seg2/pexels-#1_im.pdf}&&%
    \cropim{#1_sep1}&\cropim{#1_sep2}&&\cropim{#1_reg1}&\cropim{#1_reg2}\\%
}
\newcommand{\qualfig}[8]{\begin{figure*}[t]%
\setlength{\tabcolsep}{0.0pt}
\scriptsize
\begin{tabularx}{\textwidth}{Yp{5pt}YYp{5pt}YY}%
 && \multicolumn{2}{c}{(a) Separate skeleton prediction} && \multicolumn{2}{c}{(b) With our proposed ACAE regularization}\\
&& Front view & Right side view && Front view & Right side view \\
\figline{#1}%
\figline{#2}%
\figline{#3}%
\figline{#4}%
\figline{#5}%
\figline{#6}%

\end{tabularx}%
\caption{#7}
\label{fig:#8}
\end{figure*}
}

\qualfig{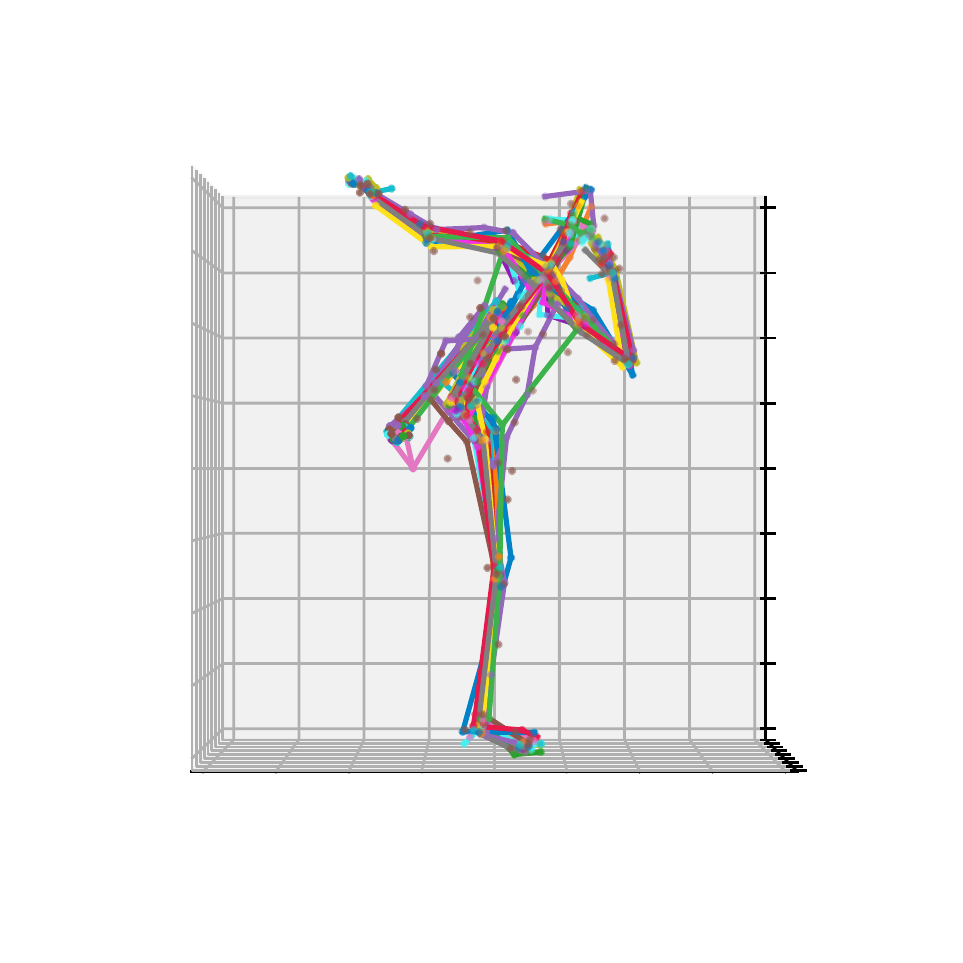}{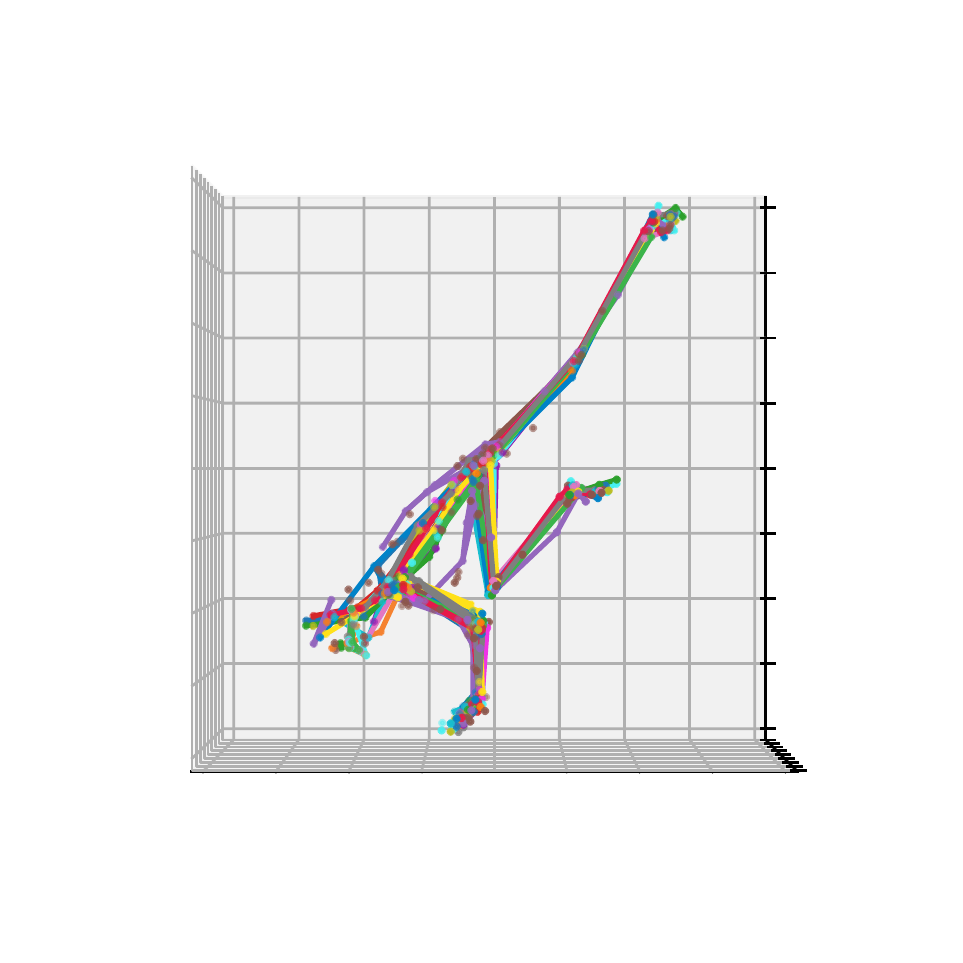}{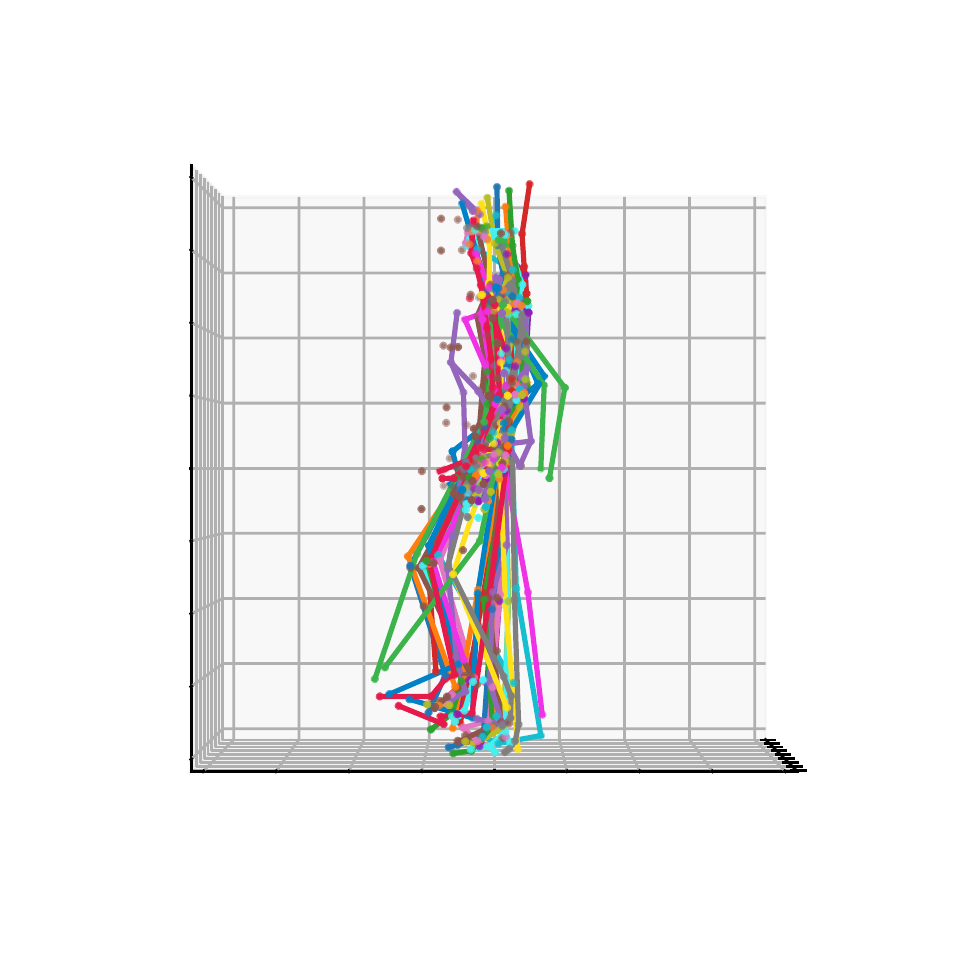}{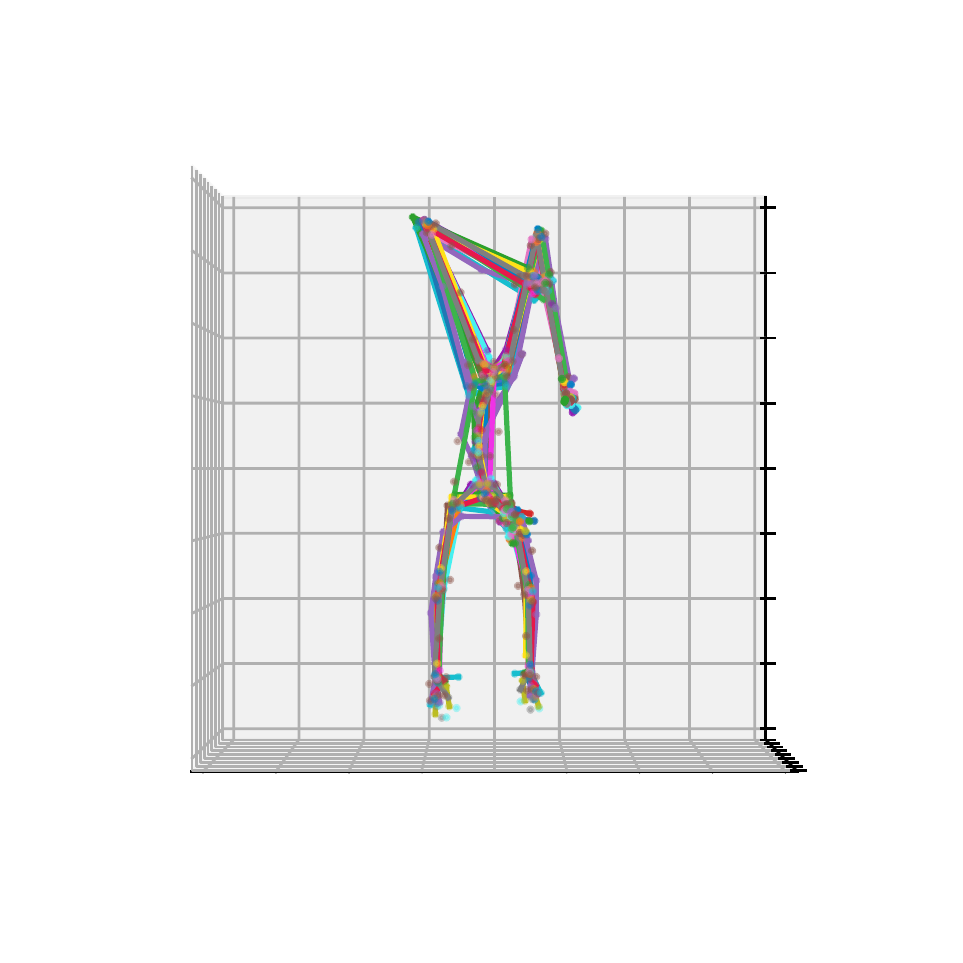}{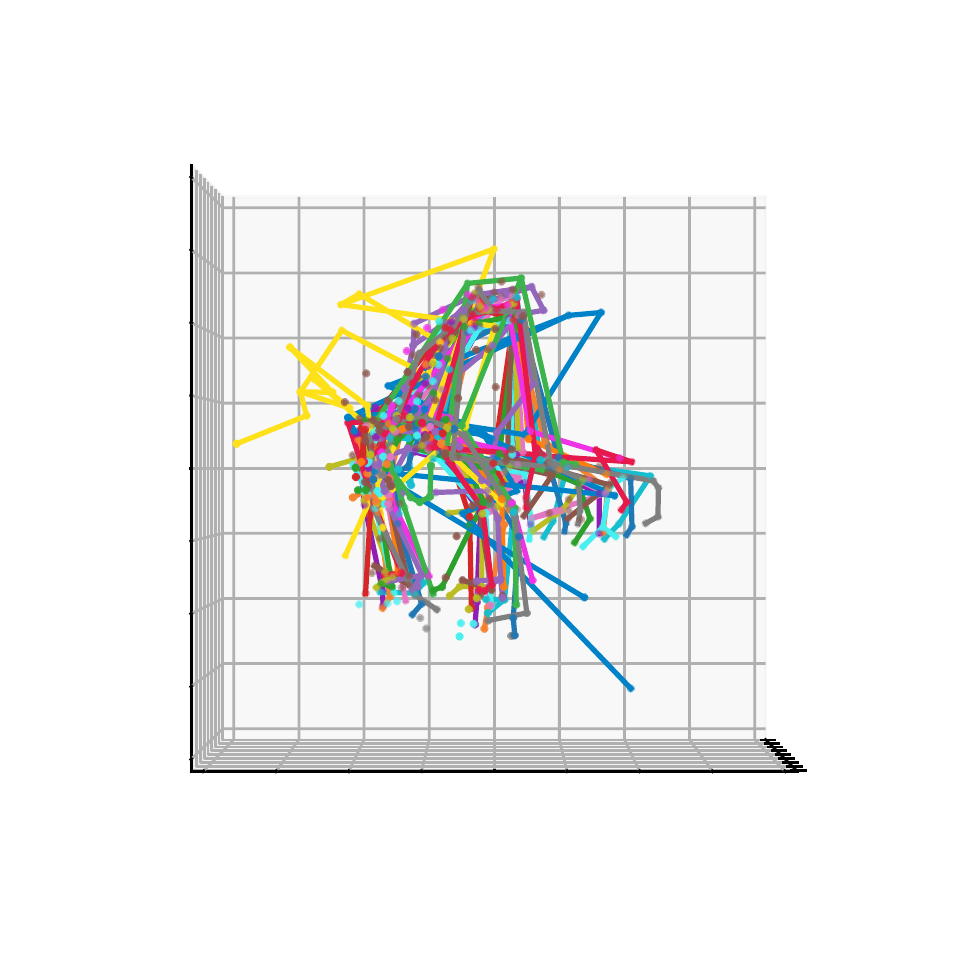}{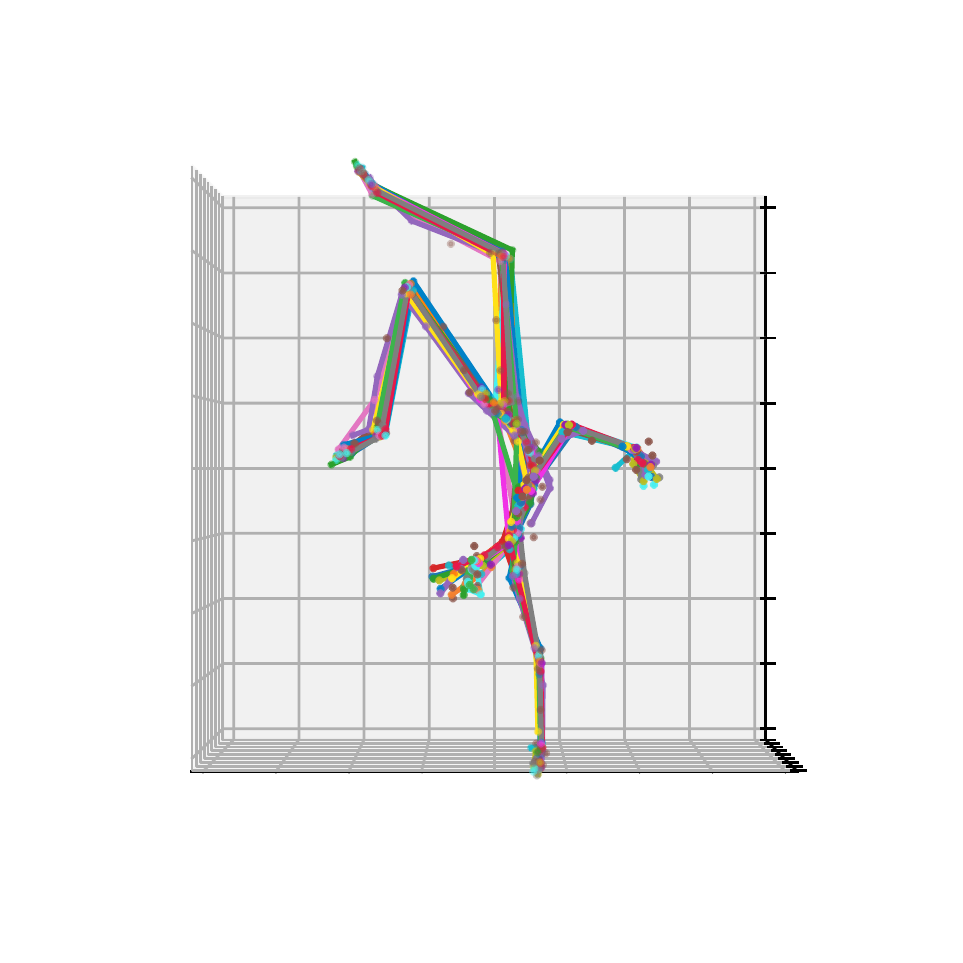}{
    A qualitative result comparison between a model trained without (a) and with our ACAE regularization (b).
    It can clearly be seen that our regularization leads to improved skeleton consistency.
}{qualitative_results1}
\qualfig{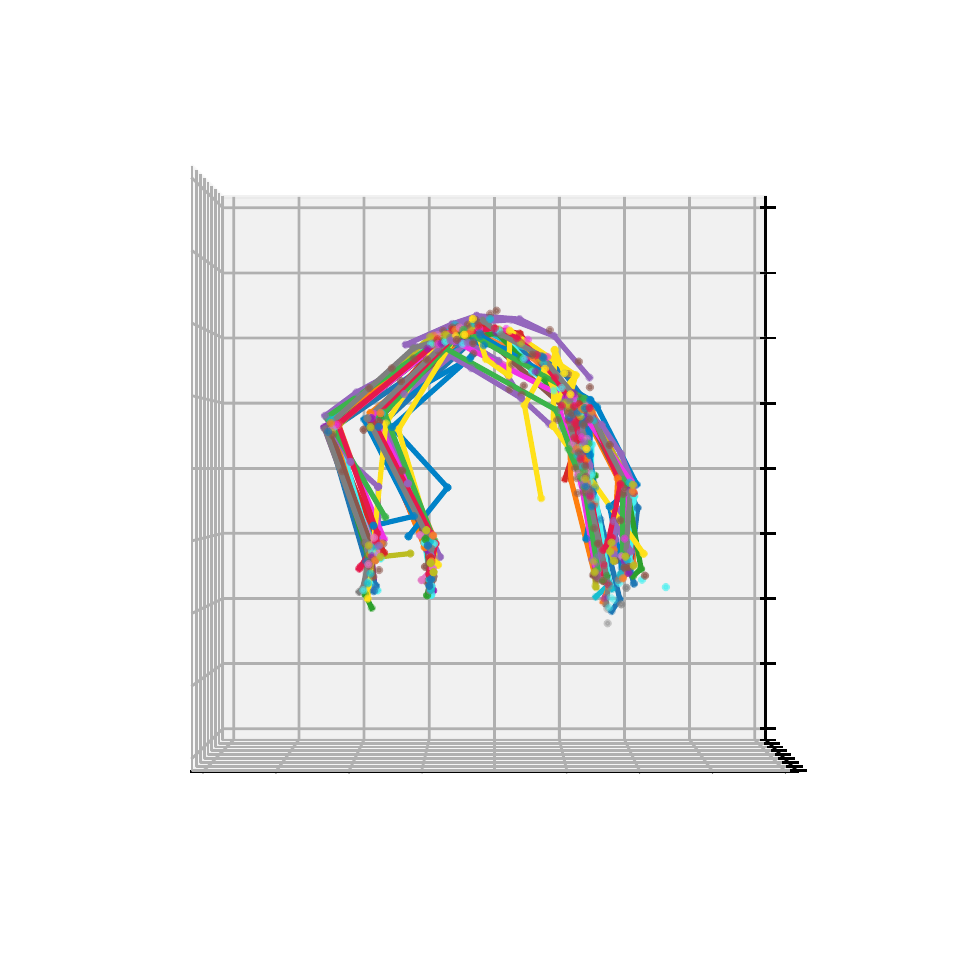}{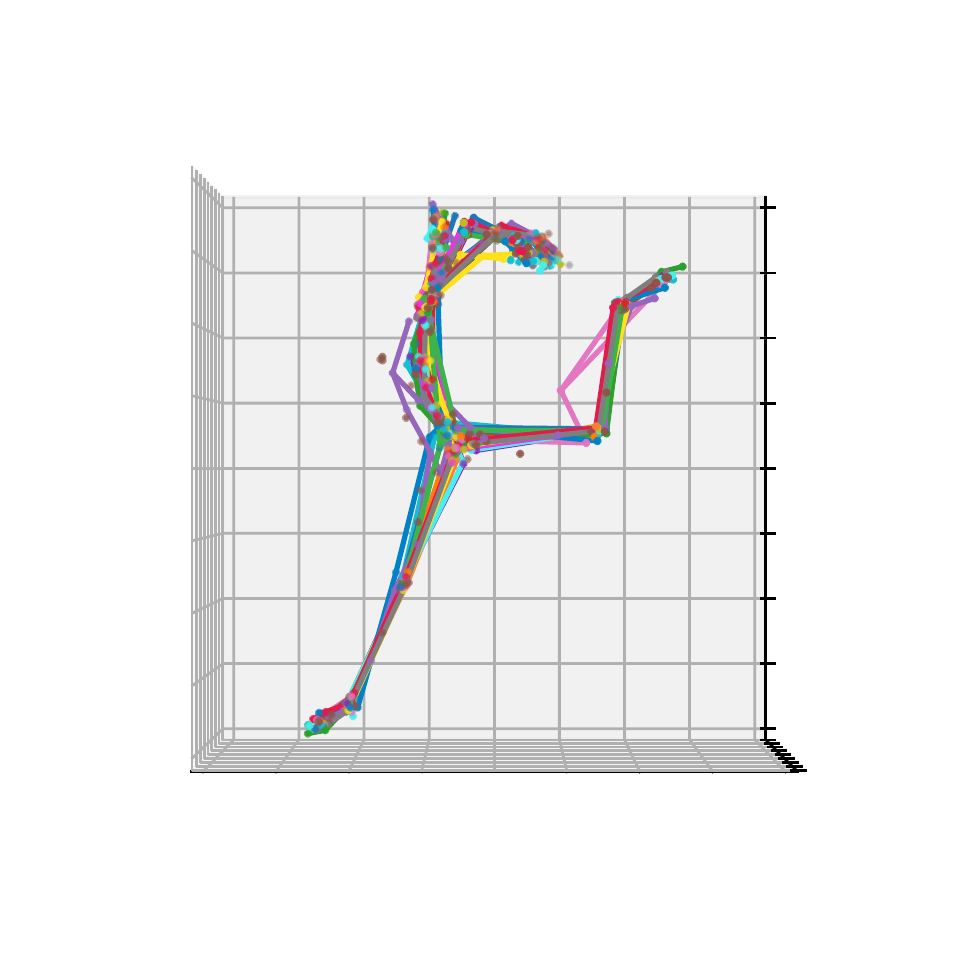}{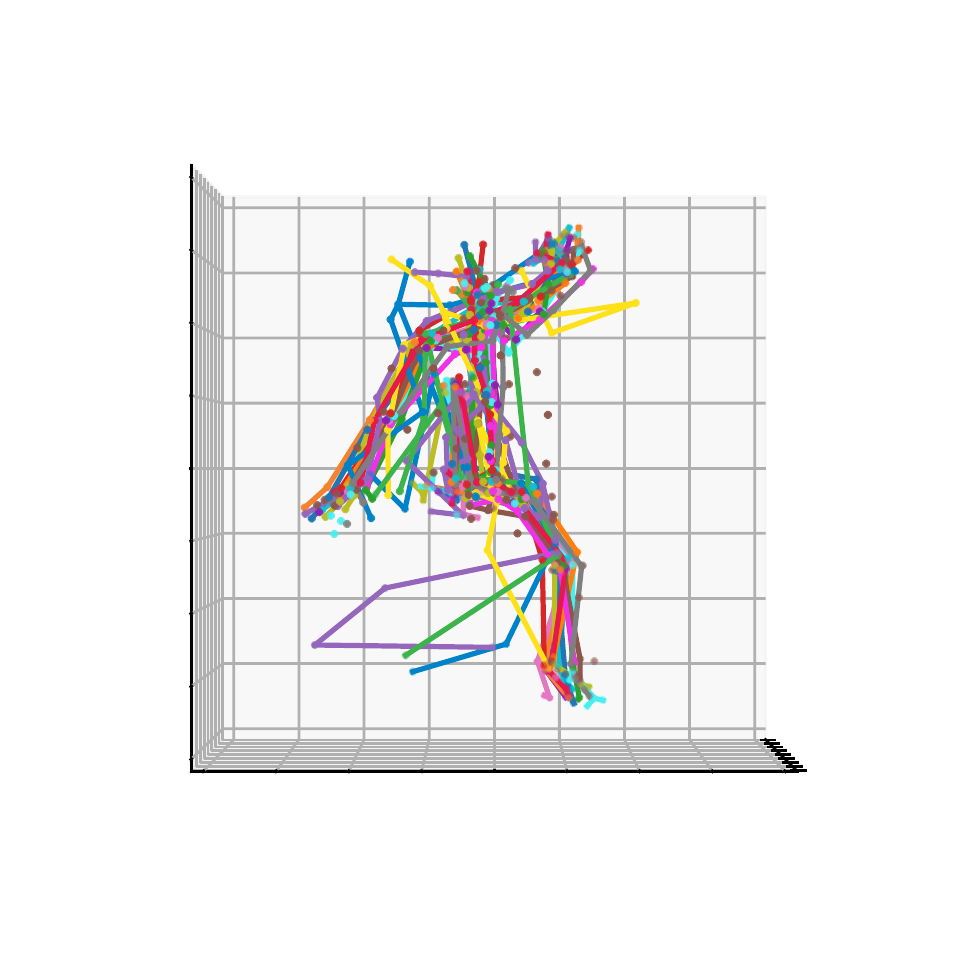}{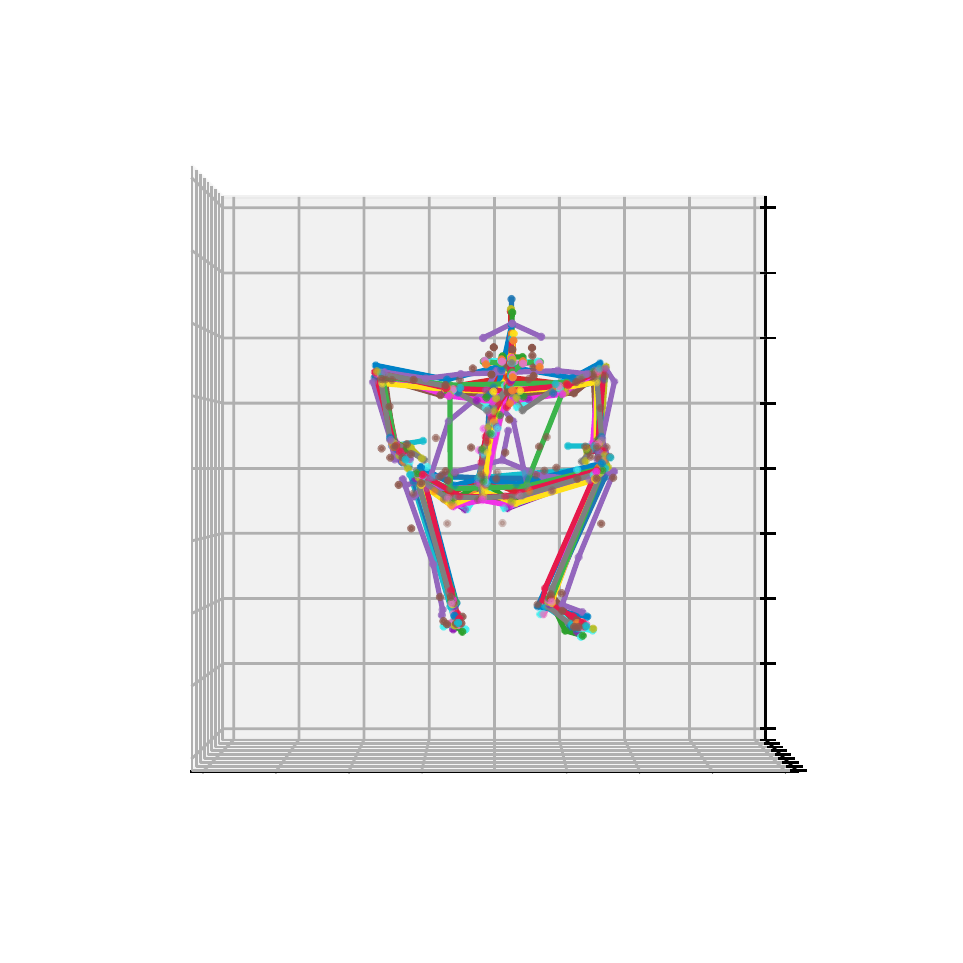}{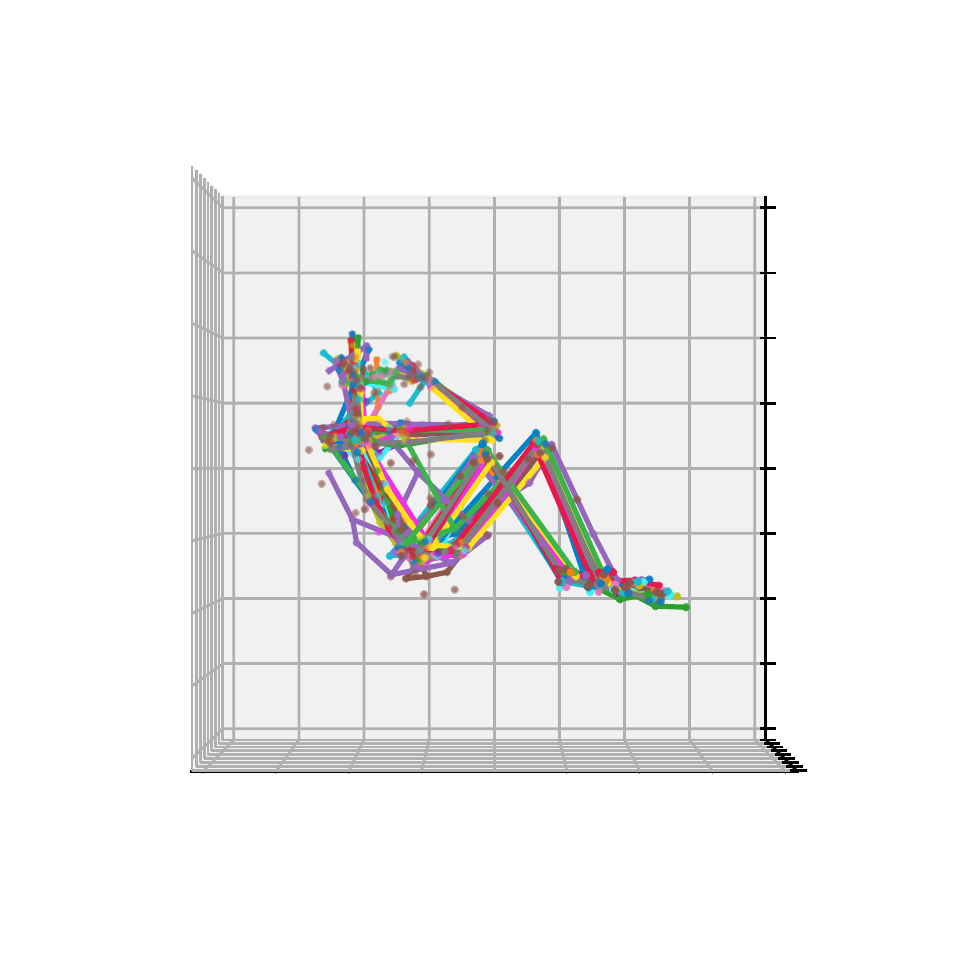}{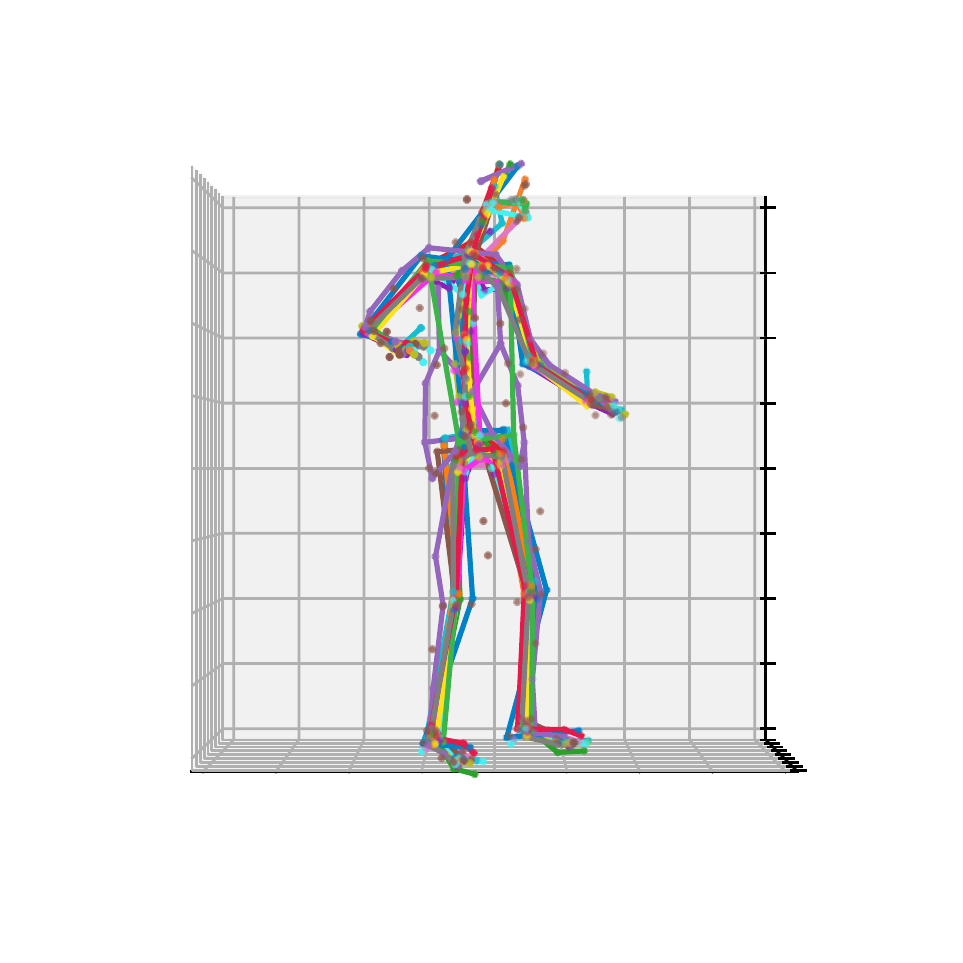}{
    A qualitative result comparison between a model trained without (a) and with our ACAE regularization (b).
    It can clearly be seen that our regularization leads to improved skeleton consistency.
}{qualitative_results2}
\qualfig{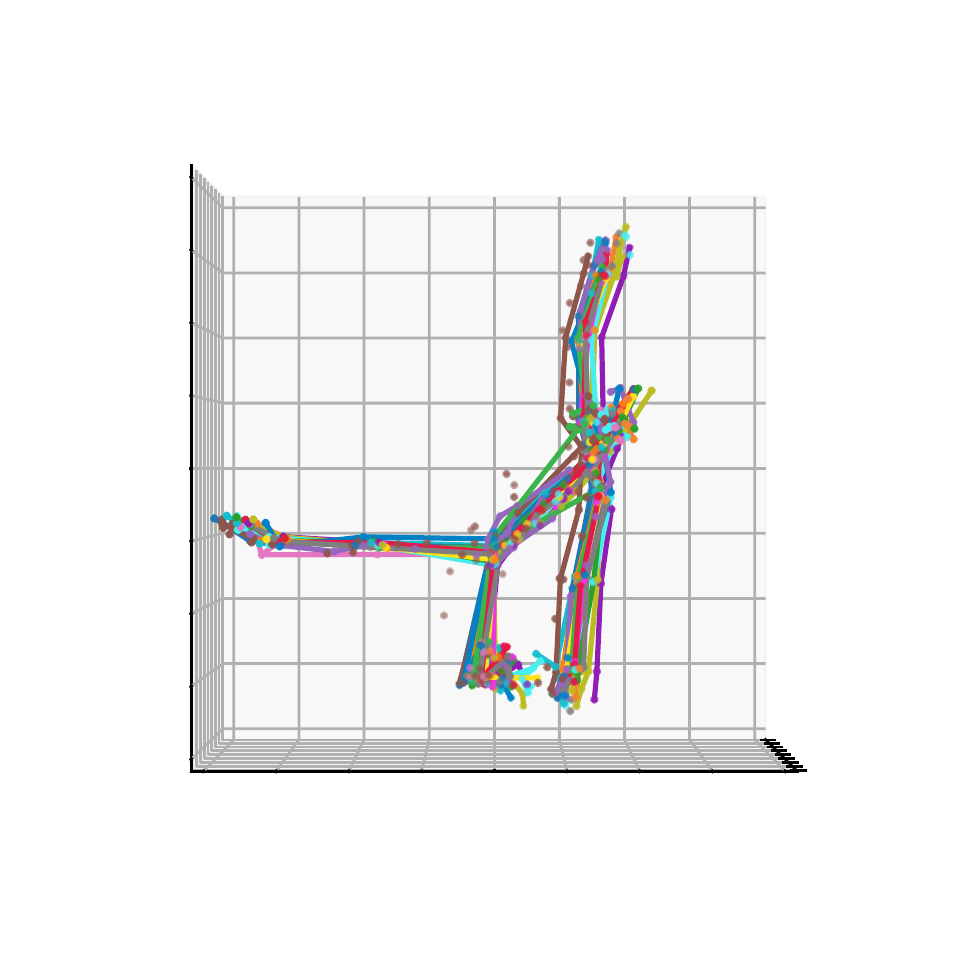}{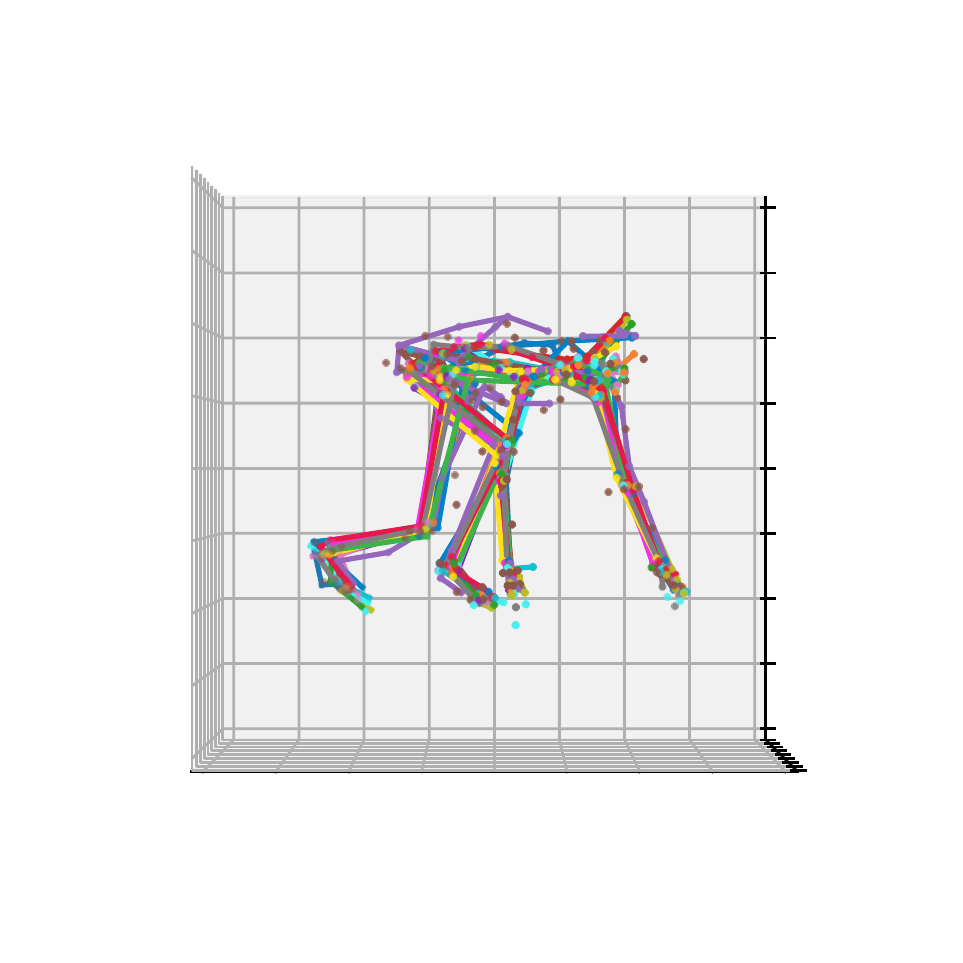}{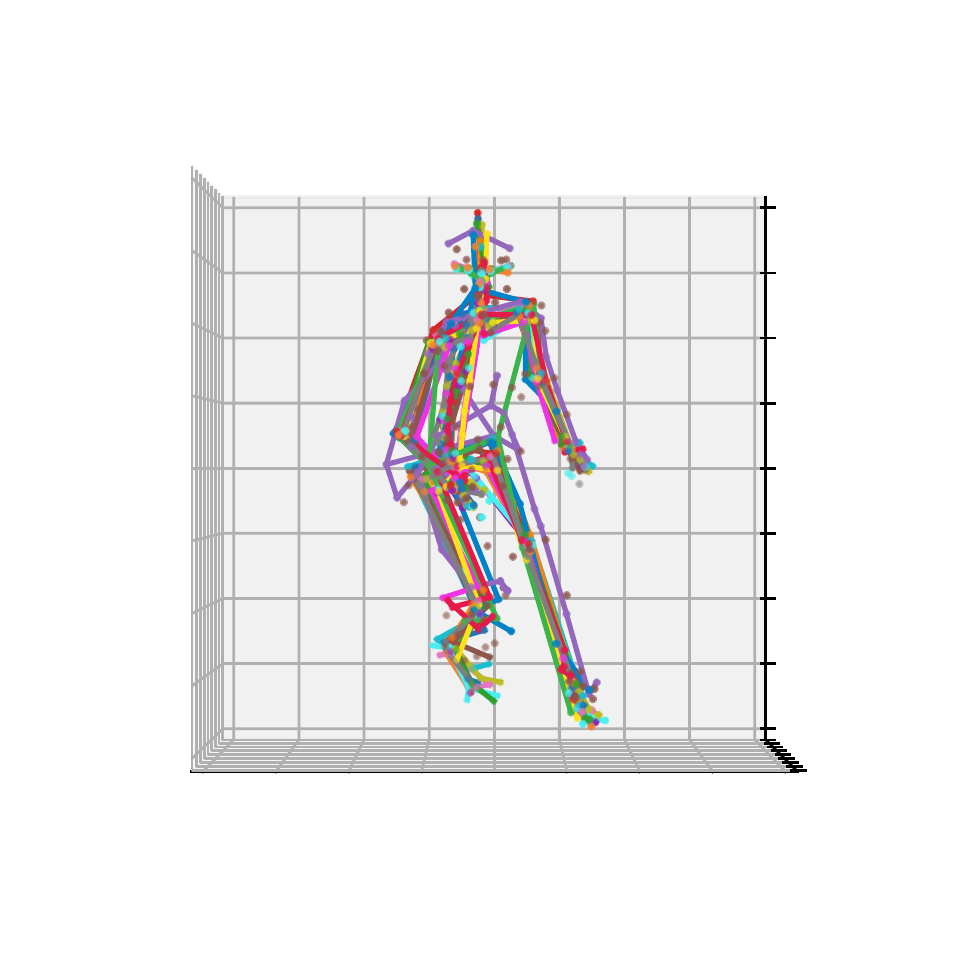}{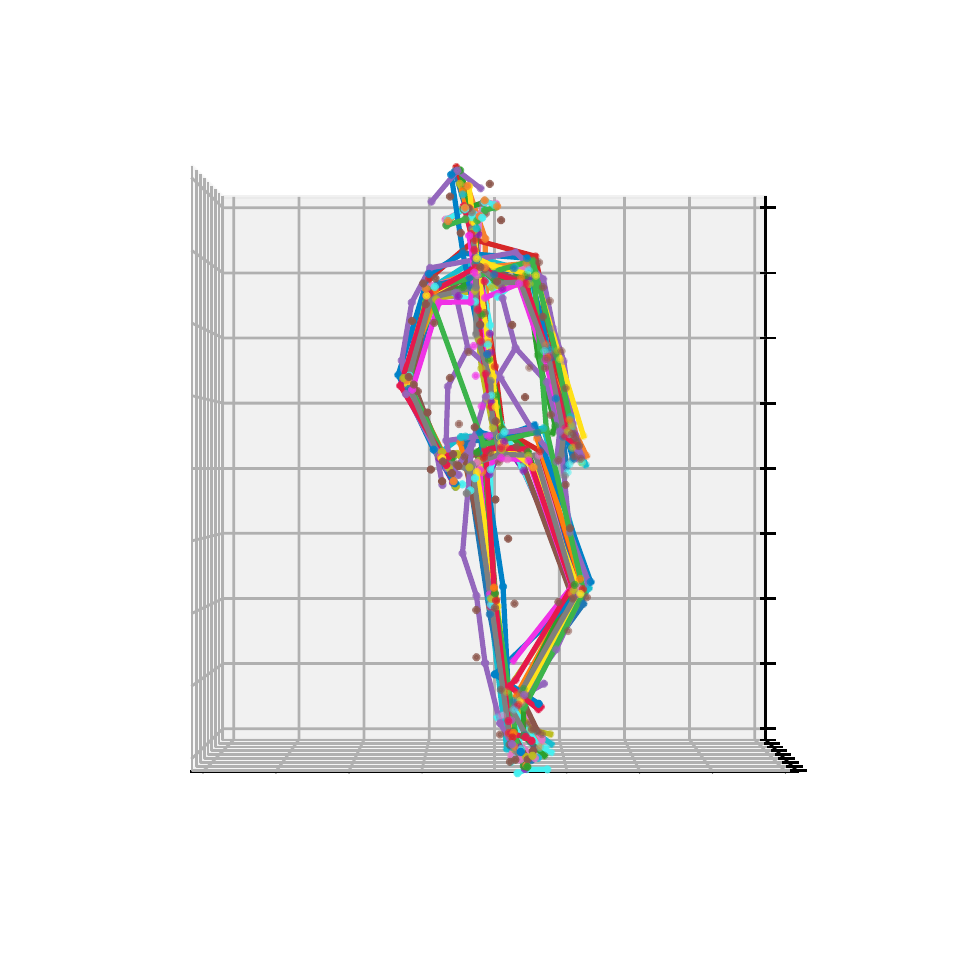}{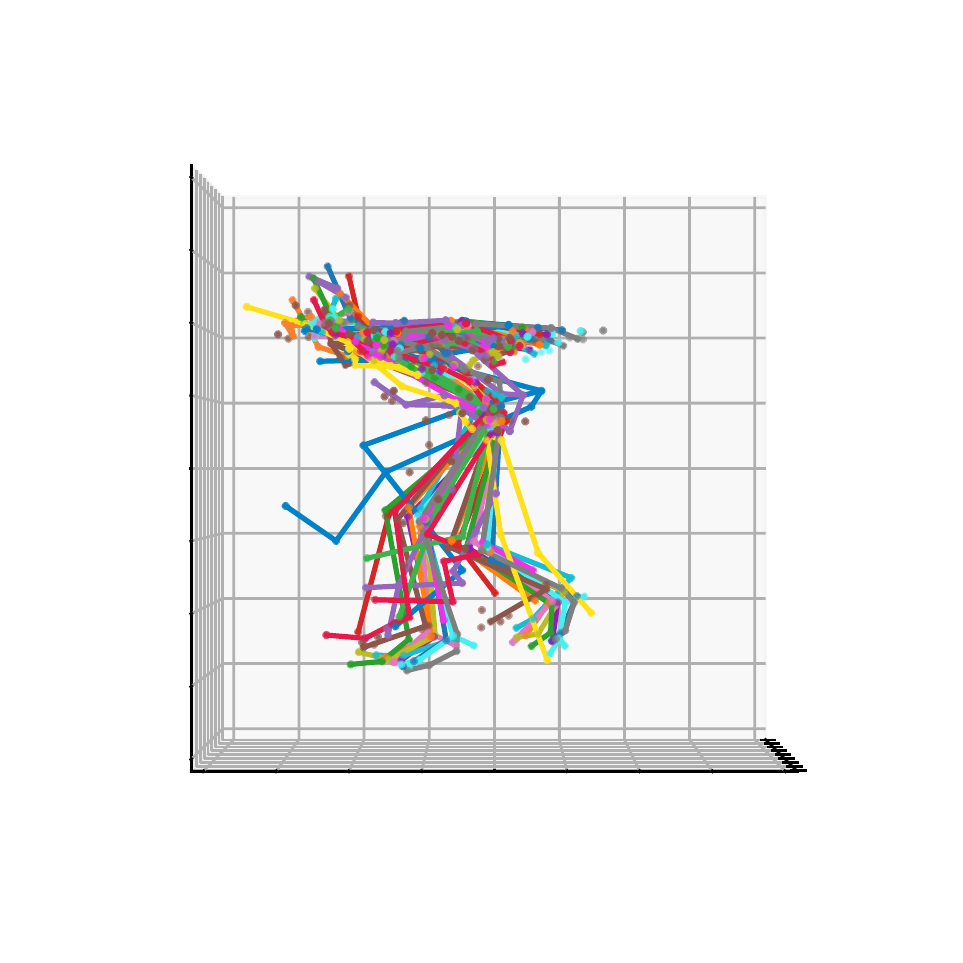}{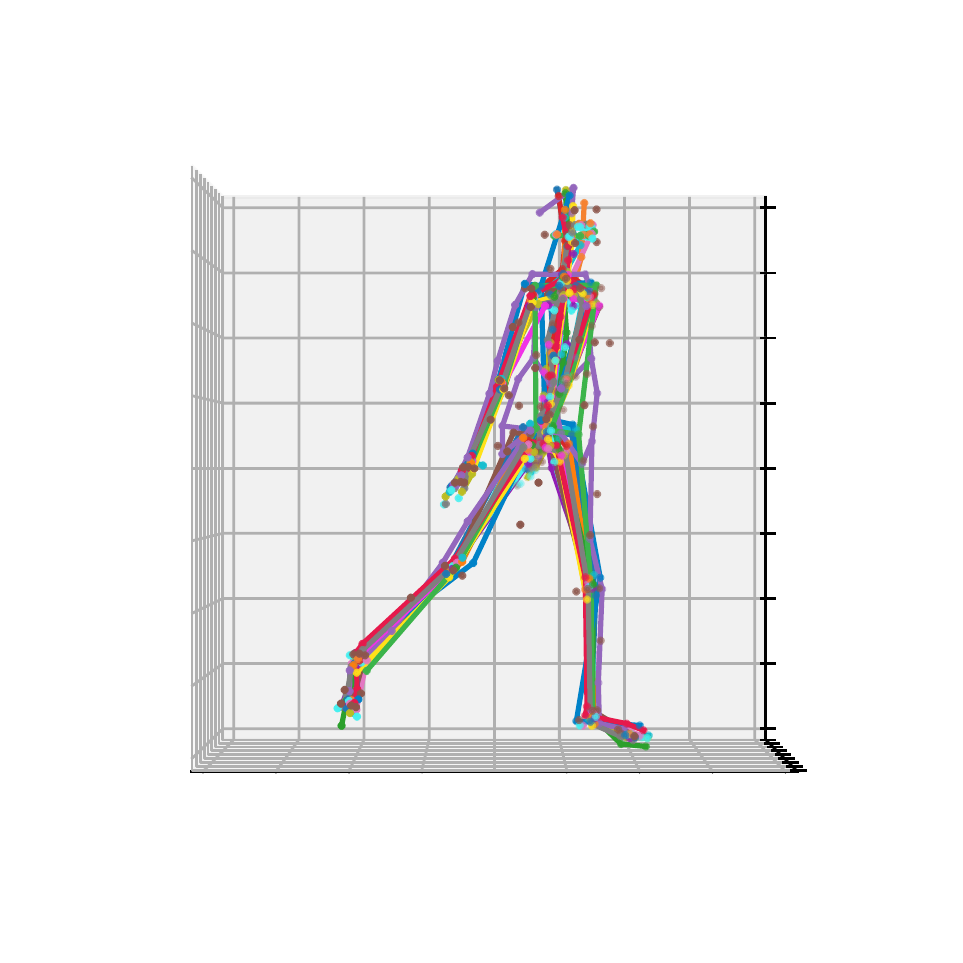}{
    A qualitative result comparison between a model trained without (a) and with our ACAE regularization (b).
    It can clearly be seen that our regularization leads to improved skeleton consistency.
}{qualitative_results3}